\documentclass[11pt]{article}
\PassOptionsToPackage{table}{xcolor}

\usepackage[preprint]{acl}

\usepackage{times}
\usepackage{latexsym}

\usepackage[T1]{fontenc}

\usepackage{microtype}

\usepackage{inconsolata}

\usepackage{graphicx}
\usepackage{enumitem}
\usepackage{amsmath}
\usepackage[ruled,vlined,linesnumbered,noend]{algorithm2e}
\DontPrintSemicolon
\usepackage{algpseudocode}
\usepackage{booktabs}
\usepackage{xcolor}
\usepackage{pgfplots}
\pgfplotsset{compat=1.18} 

\usepackage{tabularx}
\usepackage{makecell} 
\usepackage{amssymb}
\usepackage{amsthm}
\newtheorem{proposition}{Proposition}
\newtheorem{lemma}{Lemma}
\newtheorem{theorem}{Theorem}
\usepackage{placeins}
\usepackage{multirow}
\usepackage[most]{tcolorbox}
\definecolor{venuebg}{HTML}{D4ECF8}
\definecolor{modelrow}{HTML}{F7F4E8}
\definecolor{oursrow}{HTML}{F1F5F9}
\colorlet{bridgebg}{oursrow}

\usepackage{pifont}

\newtcbox{\venuebox}{on line,
  boxrule=0pt,
  colback=venuebg,
  colframe=gray!15,
  arc=1.5pt,
  left=2pt,right=2pt,top=2pt,bottom=2pt,
  boxsep=0pt}

\newcommand{\venuetag}[1]{\,{\scriptsize\venuebox{#1}}}

\definecolor{baselinecolor}{gray}{0.38}
\definecolor{resultbest}{HTML}{D0443E}
\definecolor{resultsecond}{HTML}{287FB1}
\newcommand{\best}[1]{\textcolor{resultbest}{\textbf{\boldmath #1}}}
\newcommand{\second}[1]{\textcolor{resultsecond}{\underline{\boldmath #1}}}
\title{Token Utility Is Selection-Conditioned: Coupled Selection of Prompt Context and Response Supervision for Instruction Tuning}

\author{
Can Wu 
\quad Xinrui Chen 
\quad Ou Wu\thanks{Corresponding author.} 
\quad Yi Du\\
Hangzhou Institute for Advanced Study,\\
University of Chinese Academy of Sciences, Hangzhou, China\\
\texttt{\{wucan261, chenxinrui25\}@mails.ucas.ac.cn}
}

\begin{document}
\maketitle
\begin{abstract}
Token selection for instruction tuning is coupled: supervision utility depends on retained context, and vice versa. Existing methods largely optimize prompt compression or response supervision separately. Combining selectors can value supervision using removed context, or context using excluded supervision, creating selection-state mismatch. \textbf{BRIDGE} (\textbf{\underline{B}}udgeted \textbf{\underline{R}}esponse-Prompt \textbf{\underline{I}}nteraction via \textbf{\underline{D}}irectional \textbf{\underline{G}}radient-guided \textbf{\underline{E}}fficient Token Selection) models selection-conditioned token utility through a shared validation-directed surrogate objective. BRIDGE routes response utility to supporting prompt tokens through attention, then alternates budgeted selection to revalue each side under the other's retained subset. Updates reuse precomputed interactions without repeated model scoring, while structure-aware projection promotes coherent supervision spans. Across three model families and three task domains, BRIDGE performs best overall among evaluated selection methods. Its MATH advantage over independent selection grows with compression, reaching 1.66 percentage points on Llama-3.2-3B. \href{https://anonymous.4open.science/r/BRIDGE-E8CA/}{Code}.
\end{abstract}

\section{Introduction}

\begin{figure}[t]
    \centering
    \includegraphics[width=\linewidth]{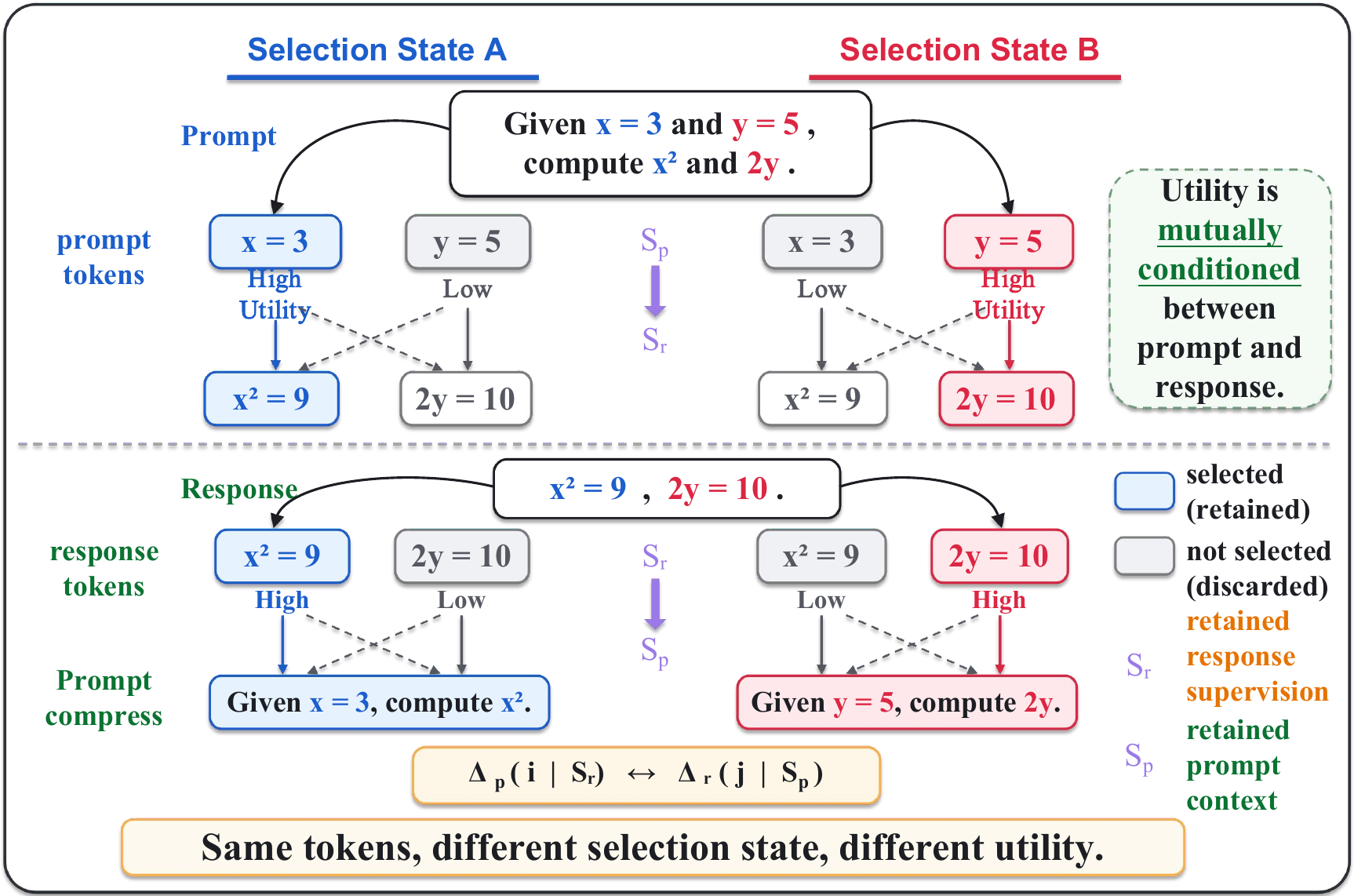}
    \vspace{-0.3in}
    \caption{Token utility depends on retained context and supervision across prompt--response selection states. Response selection controls loss positions; the complete response remains available as causal history.}
    \label{fig:teaser_selection_conditioned}
    \vspace{-0.2in}
\end{figure}

Instruction tuning adapts large language models (LLMs) using prompt--response pairs~\citep{wei2022finetuned,ouyang2022training,pareja2025secret}: prompts provide context, while responses provide learning targets. Under token budgets, effective selection must coordinate what the model sees with where it receives supervision. Consider the example in Fig.~\ref{fig:teaser_selection_conditioned}: supervising $x^2=9$ makes the prompt fact $x=3$ directly relevant; shifting supervision to $2y=10$ changes which fact matters. Conversely, removing $x=3$ changes the contextual support for supervising $x^2=9$. \textit{\textbf{Token utility is therefore selection-conditioned: each side's value depends on the retained subset of the other.}}

Existing approaches largely separate these roles. Prompt compression retains informative context~\citep{jiang2024longllmlingua,pan2024llmlingua}, response-oriented SFT filters or reweights supervision~\citep{pang2025token,qin2026sstoken,liu2026profit}, and training-time sparsification targets activation and backpropagation costs~\citep{simoulin2024memory,zeng2026tokenseek}. Combining prompt and response selectors can nevertheless value supervision using context that is later removed, or context using supervision that is later excluded. We call this discrepancy selection-state mismatch. Selecting the prompt first introduces only one-way conditioning: response selection observes the compressed context, but prompt selection is not revisited after supervision changes. \textit{\textbf{The missing link is mutual conditioning: each selection must account for what is retained on the other side.}}

We introduce \textbf{BRIDGE}, which jointly selects prompt context and response supervision under a shared validation-directed surrogate objective. Gradient alignment estimates response-token utility for the target task, and response-to-prompt attention routes that utility to supporting context. Both sides' conditional scores are marginal gains of the same objective. Budgeted alternating updates therefore revalue each subset against the other's current selection while optimizing a common criterion. Reaggregating precomputed interactions enables this coordination without repeated model scoring.

Because token-level selection can fragment useful supervision into isolated positions, a structure-aware projection favors coherent expressions, reasoning steps, and code statements within the response budget. Final SFT restarts from the pretrained model, executes the compacted prompt, and retains the complete response as causal history while applying loss only at selected positions.

Across mathematical reasoning, code generation, and instruction following, BRIDGE achieves the highest mean target-task performance and average general capability among the evaluated selection methods on every backbone. Relative to pretrained baselines, average general capability increases in eight of nine configurations and decreases by $0.40$ percentage points for Qwen code fine-tuning. With scoring, budgets, projection, and final training fixed, the MATH gain of BRIDGE over independent selection on Llama-3.2-3B increases from $0.34$ to $1.66$ percentage points as prompt and response retention decrease from $90\%$ to $50\%$, demonstrating larger gains from coordinated context and supervision selection under stronger compression.

Our contributions are threefold:
\begin{itemize}[leftmargin=*, nosep]
    \item We provide a theoretical and empirical analysis that exposes selection-state mismatch in independent token selection and validates budgeted coordination of prompt context and response supervision as an effective remedy.
    \item We develop BRIDGE with a shared validation-directed surrogate for both sides' conditional utilities, budgeted alternating selection over precomputed interactions, and structure-aware projection for coherent supervision.
    \item We demonstrate the best overall performance among evaluated selection methods across three model families and three task domains, while preserving average general capability at or above pretrained levels in mathematical reasoning.
\end{itemize}

\section{Related Work}

\textbf{Prompt-Side Reduction.} Prompt-side methods reduce input length or training computation. Compression uses information-theoretic or perplexity-based scores~\citep{li2023compressing,jiang2023llmlingua}, query-aware relevance~\citep{jiang2024longllmlingua}, or distilled token classifiers~\citep{pan2024llmlingua}. Training-time sparsification reduces activation memory and backward computation through restricted token-level gradient propagation or selective token dropping~\citep{simoulin2024memory,zeng2026tokenseek}.

\textbf{Response-Side Selection.} Response-side methods filter supervision using token probabilities and loss changes~\citep{pang2025token,liu2026profit}, alongside attention-based importance, reasoning relevance, and knowledge novelty~\citep{qin2026sstoken,yang2026explainable}. Objective-based valuation combines local gradients and downstream autoregressive influence to balance task adaptation and capability retention~\citep{liu2026alphatoken}. Token informativeness guides sample selection through example-level quality assessment~\citep{fu2025tshirt}. Loss-weighting studies examine how prompt and response contributions affect generalization~\citep{huerta2024instruction,chatterjee2025effect}. BRIDGE conditions context and supervision selection on each other's retained subset.

\textbf{Data Valuation and Selection Granularity.} Example-level valuation uses quality, diversity, or difficulty proxies~\citep{liu2024what,li2024superfiltering}, or influence and gradient alignment to estimate target contributions~\citep{koh2017understanding,xia2024less,min2026gist}, with scoring-cost trade-offs~\citep{yin2025computeconstrained}. Joint sample--token selection captures utility variation across and within examples~\citep{chen2025quadratic,wang2025winning}. Attention offers interaction proxies without establishing causal attribution~\citep{jain2019attention,serrano2019attention}.

\textbf{Capability Retention during Fine-Tuning.} Retention methods adjust learning signals through example reweighting or difficult/high-perplexity token suppression~\citep{sanyal2025upweighting,lin2026sft,wu2025mitigating}, or control parameter updates through flatter solutions, selective updates, or task-vector pruning~\citep{li2024revisiting,zhou2025payattentionsmallweights,huang2025mitigating}. BRIDGE couples context--supervision selection without additional retention regularization during final SFT.

\section{Token Utility Is Selection-Conditioned}
\label{sec:coupling_analysis}

\begin{figure}[t]
\centering
\includegraphics[width=\columnwidth]{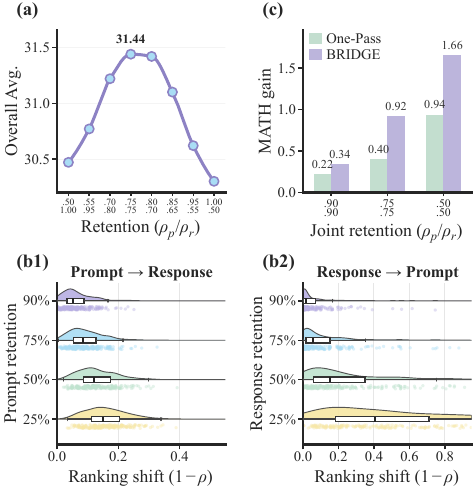}
\vspace{-0.3in}
\caption{Evidence for coupled selection: (a) allocation, (b) ranking shifts, and (c) gains under compression.}
\label{fig:coupling_analysis}
\vspace{-0.1in}
\end{figure}

Token utility depends on retained context and supervision. We characterize when full-state selection fails and demonstrate the benefits of coordinating both sides under compression.

\textbf{\ding{228} Finding 1: Joint allocation outperforms one-sided reduction.} With $\rho_p+\rho_r=1.5$, retaining $75\%$ on each side outperforms both tested one-sided alternatives (Fig.~\ref{fig:coupling_analysis}(a)), motivating selection across both token roles.

\textbf{\ding{228} Finding 2: Selection changes token rankings.} Fig.~\ref{fig:coupling_analysis}(b) shows larger ranking shifts, measured as one minus Spearman correlation with full-retention rankings, as opposite-side retention decreases. Both prompt and response rankings exhibit this dependency on opposite-side retention. Response reduction masks supervision while preserving causal history. Thus, full-state rankings can become misaligned with retained states.

\textbf{Selection-state robustness.} Ranking shifts invalidate selection when they cross its top-$b$ boundary. For eligible current-side positions $\mathcal{X}$ and opposite-side positions $\mathcal{Y}$, consider BRIDGE's fixed additive scores (Section~\ref{sec:method}):
\begin{equation}
s_i(V)=a_i+\sum\nolimits_{j\in V}B_{ij}, \qquad V\subseteq\mathcal{Y}.
\label{eq:additive_conditional_score}
\end{equation}
Let $n=|\mathcal{Y}|$ and $S^F$ maximize $U_V(S)=\sum_{i\in S}s_i(V)$ over size-$b$ subsets at $V=\mathcal{Y}$, with $1\leq b<|\mathcal{X}|$. For each $i\in S^F$ and $i'\in\mathcal{X}\setminus S^F$, sort $d_{ii',j}=B_{ij}-B_{i'j}$ increasingly. For an opposite-side budget $1\leq q\leq n$, define the worst-case boundary margin:
\begin{equation}
\kappa_q=\min_{\substack{i\in S^F\\ i'\in\mathcal{X}\setminus S^F}}\left[a_i-a_{i'}+\sum_{1\leq\ell\leq q}d_{ii',(\ell)}\right].
\label{eq:kappa_main}
\end{equation}

\begin{theorem}
\label{thm:selection_state_robustness}
Before structure-aware projection, $S^F$ maximizes $U_V(S)$ over size-$b$ selections for every size-$q$ opposite-side subset if and only if $\kappa_q\geq0$. If $\kappa_q<0$, a minimizing pair and its $q$ smallest interaction differences identify a subset admitting a strictly improving swap.
\end{theorem}

This criterion identifies exactly when full-state selection remains optimal across every opposite-side state of a given size. For prompts ($a_i=0$), $\kappa_q/q$ is nondecreasing in $q$ for fixed $b$, $S^F$, and $B$: stronger response compression cannot improve worst-case normalized margins. Appendix~\ref{app:selection_state_theory} provides the proof and recovered-utility expression.

\textbf{\ding{228} Finding 3: Coordination gains grow with compression.} Fig.~\ref{fig:coupling_analysis}(c) compares Independent (full-state scoring), One-Pass ($T=1$), and BRIDGE ($T=4$), matching budgets, precomputed interactions, history utilities, projection, and final training; only schedules differ. BRIDGE's MATH gains over Independent on Llama-3.2-3B rise from $0.34$ to $0.92$ and $1.66$ percentage points as both retention ratios fall from $90\%$ to $75\%$ and $50\%$. BRIDGE exceeds One-Pass across tested budgets, demonstrating benefits beyond one round. Appendix~\ref{app:coupling_definitions} defines schedules.

These findings connect selection-state mismatch to improved training through coordinated revaluation, motivating BRIDGE's shared surrogate and budgeted alternating selection.

\section{Methodology}
\label{sec:method}

\begin{figure*}[t]
\centering
\includegraphics[width=\linewidth]{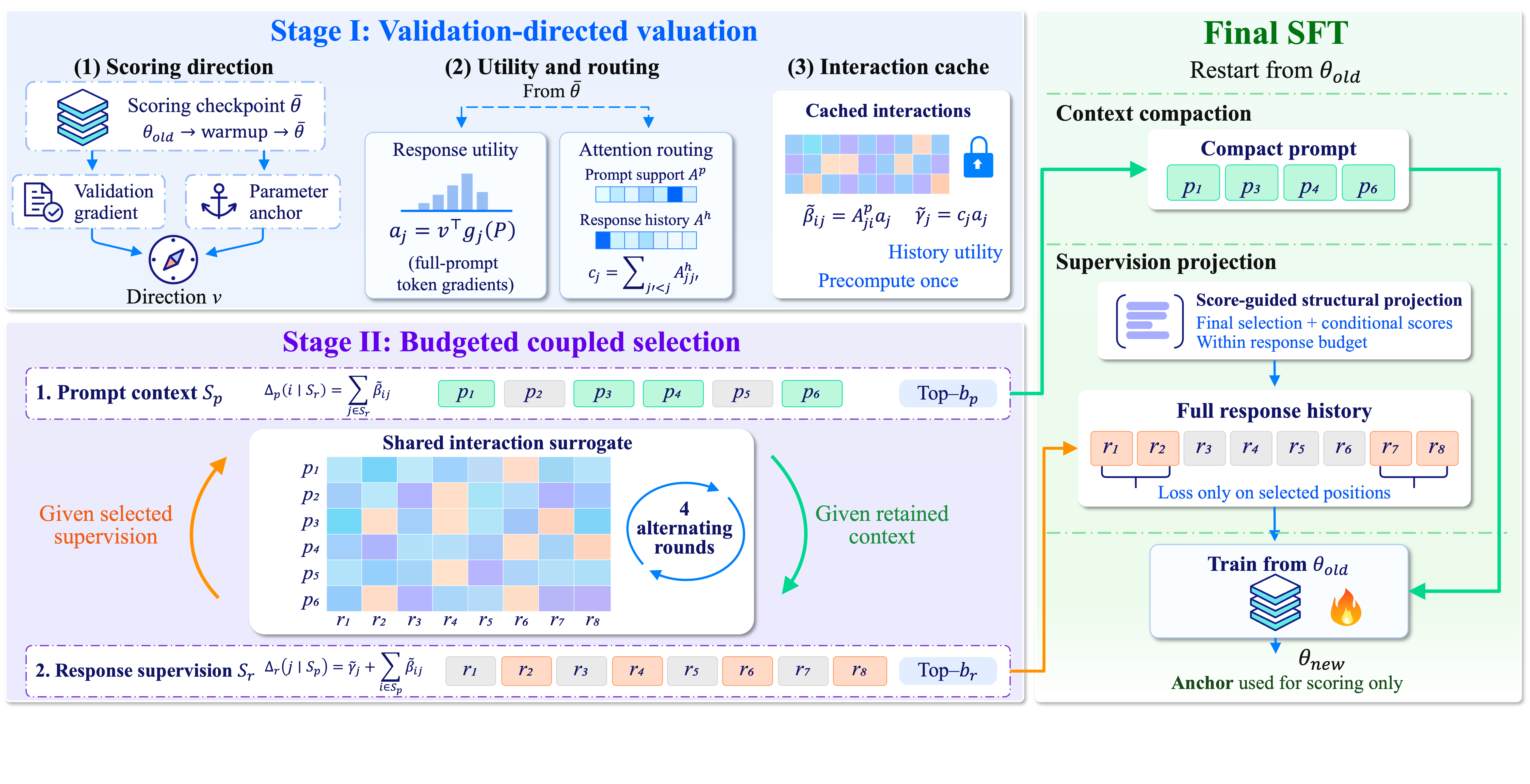}
\vspace{-0.6in}
\caption{Overview of BRIDGE: Coupled Prompt-Response Valuation for Instruction Tuning.}
\label{fig:overview}
\vspace{-0.1in}
\end{figure*}

BRIDGE jointly selects prompt context and response supervision through validation-directed token valuation, budgeted alternating selection, and structural projection.

\subsection{Problem Setup}
\label{subsec:coupled_objective}

Prompts provide context, while responses provide supervision. Let $D=\{(P_k,R_k)\}_{k=1}^{N}$ pair prompts $P_k=(p_{k,1},\ldots,p_{k,m_k})$ and responses $R_k=(r_{k,1},\ldots,r_{k,n_k})$, with eligible token position sets $\mathcal{P}_k$ and $\mathcal{R}_k$. BRIDGE selects $S_{p,k}\subseteq\mathcal{P}_k$ and $S_{r,k}\subseteq\mathcal{R}_k$, preserving order. Fixed boundary positions $F_{p,k}$ and $F_{r,k}$ are disjoint from eligible positions and excluded from selection budgets. For per-example budgets $b_{p,k}$ and $b_{r,k}$, the alternating stage selects $\mathcal{S}=\{(S_{p,k},S_{r,k})\}_{k=1}^{N}$ with $|S_{p,k}|=b_{p,k}$ and $|S_{r,k}|=b_{r,k}$; structural projection enforces the response-budget cap. Ideal selection minimizes downstream validation loss under token constraints~\citep{wu2026computational}:
\begin{equation}
\label{eq:ideal_selection}
\min\nolimits_{\mathcal{S}}\; L_{\mathrm{val}}\!\left(\operatorname{Train}_{\mathcal{A}}(\theta_{\mathrm{old}};D,\mathcal{S})\right).
\end{equation}
Here, $\operatorname{Train}_{\mathcal{A}}$ denotes training from $\theta_{\mathrm{old}}$ under training scheme $\mathcal{A}$. Context and supervision jointly determine the training signal, so their selections generally cannot be optimized independently.

\subsection{Validation-Directed Token Valuation}
\label{subsec:token_valuation}

BRIDGE uses a local scoring surrogate to avoid retraining for each candidate selection or differentiating through training trajectories~\citep{grazzi2020hypergradient}. A standard SFT warmup yields the scoring checkpoint $\bar{\theta}$. The warmup split $D_{\mathrm{w}}$, validation split $D_{\mathrm{val}}$, and selection/training split $D$ are pairwise disjoint (Appendix~\ref{app:data_split}).

Let $\mathcal{G}$ denote the scoring parameter subset. We compute the validation gradient $g_{\mathrm{target}}=\nabla_{\theta_{\mathcal{G}}}L_{\mathrm{val}}(\bar{\theta};D_{\mathrm{val}})$ and anchor direction $g_{\mathrm{anchor}}=\bar{\theta}_{\mathcal{G}}-\theta_{\mathrm{old},\mathcal{G}}$. The anchor is used only during scoring to favor updates that counteract parameter drift. Normalizing each component separately yields
\begin{equation}
v=\frac{(1-\lambda)g_{\mathrm{target}}}{\|g_{\mathrm{target}}\|_2+\epsilon}
+\frac{\lambda g_{\mathrm{anchor}}}{\|g_{\mathrm{anchor}}\|_2+\epsilon},
\end{equation}
where $\lambda$ balances the two components and $\epsilon>0$ ensures numerical stability. Positive gradient alignment estimates a decrease in a weighted sum of validation loss and squared parameter drift under a first-order gradient-descent approximation. Scoring configurations and the gradient interpretation are given in Appendices~\ref{app:scoring_config} and~\ref{app:local_descent}.

For a prompt $Q$, let $g_{k,j}(Q)$ denote the gradient of the negative log-likelihood of $r_{k,j}$ given $(Q,R_{k,<j})$ with respect to $\theta_{\mathcal{G}}$, evaluated at $\bar{\theta}$. Define $\widetilde{P}_k=\operatorname{Compact}(P_k,F_{p,k}\cup S_{p,k})$. The local selection utility is
\begin{equation}
\label{eq:local_utility}
U_{\mathrm{loc}}(\mathcal{S})=
\sum\nolimits_{k=1}^{N}\sum\nolimits_{j\in S_{r,k}}
v^\top g_{k,j}(\widetilde{P}_k).
\end{equation}
For the full prompt, $a_{k,j}=v^\top g_{k,j}(P_k)$ is computed as a directional derivative at the scoring checkpoint without explicitly materializing per-token parameter-gradient vectors.

\subsection{Attention-Routed Interaction Surrogate}
\label{subsec:interaction_valuation}

BRIDGE approximates conditional utilities by routing full-prompt response utilities through fixed attention weights. Let $A^p_{k,j,i}$ and $A^h_{k,j,j'}$ denote routing weights from response position $j$ to prompt position $i$ and response-prefix position $j'<j$, respectively. Routing uses the query row at the input position of $r_{k,j}$, applies softmax within each head over all unmasked causal keys, and averages the weights across heads. Prompt and response-prefix weights are extracted without renormalization.

The prompt interaction and history terms are
\begin{equation}
\label{eq:interaction_surrogate}
\widetilde{\beta}_{k,i,j}=A^p_{k,j,i}a_{k,j},
\qquad
\widetilde{\gamma}_{k,j}=a_{k,j}c_{k,j},
\end{equation}
where $c_{k,j}=\sum\nolimits_{j'<j}A^h_{k,j,j'}$ includes all response-prefix positions, irrespective of supervision, and excludes self-attention. Attention supplies routing weights, while gradient alignment supplies signed utility. The coefficients remain fixed during selection, so evaluating a retained subset requires only summing the corresponding terms.

Prompt and response selection optimize the shared surrogate
\begin{equation}
\label{eq:surrogate_selection_objective}
\begin{aligned}
U_{\mathrm{sur}}(S_p,S_r)
&=\sum\nolimits_{k=1}^{N}\sum\nolimits_{j\in S_{r,k}}
\Bigl[\widetilde{\gamma}_{k,j}\\
&\qquad+\sum\nolimits_{i\in S_{p,k}}\widetilde{\beta}_{k,i,j}\Bigr],
\end{aligned}
\end{equation}
where $S_p=(S_{p,k})_{k=1}^{N}$ and $S_r=(S_{r,k})_{k=1}^{N}$. Its marginal gains define the conditional scores:
\begin{equation}
\label{eq:conditional_scores}
\begin{aligned}
\Delta_{p,k}(i\mid S_{r,k})&=\sum\nolimits_{j\in S_{r,k}}\widetilde{\beta}_{k,i,j},\\
\Delta_{r,k}(j\mid S_{p,k})&=\widetilde{\gamma}_{k,j}
+\sum\nolimits_{i\in S_{p,k}}\widetilde{\beta}_{k,i,j}.
\end{aligned}
\end{equation}
Prompt scores aggregate utility from selected supervision, while response scores combine retained prompt support with the fixed history term. Perturbation-based diagnostic references are defined in Appendix~\ref{app:exact_interaction_measurement}.

\subsection{Budgeted Coupled Selection and Structural Projection}
\label{subsec:coupled_selection}

For retention ratios $0<\rho_p,\rho_r\leq1$, set $b_{p,k}=\lceil\rho_p|\mathcal{P}_k|\rceil$ and $b_{r,k}=\lceil\rho_r|\mathcal{R}_k|\rceil$. BRIDGE initializes $S_{r,k}^{(0)}=\operatorname{Top}_{b_{r,k}}\{\widetilde{\gamma}_{k,j}\}_{j\in\mathcal{R}_k}$ and alternates
\begin{equation}
\label{eq:alternating_updates}
\begin{aligned}
S_{p,k}^{(t+1)}&=\operatorname{Top}_{b_{p,k}}
\{\Delta_{p,k}(i\mid S_{r,k}^{(t)})\}_{i\in\mathcal{P}_k},\\
S_{r,k}^{(t+1)}&=\operatorname{Top}_{b_{r,k}}
\{\Delta_{r,k}(j\mid S_{p,k}^{(t+1)})\}_{j\in\mathcal{R}_k}.
\end{aligned}
\end{equation}
Here, $\operatorname{Top}_b$ returns the indices of the $b$ largest signed scores. With fixed interaction coefficients, each alternating update exactly maximizes $U_{\mathrm{sur}}$ conditional on the other subset, yielding nondecreasing utility across feasible alternating iterates.

To promote coherent supervision, BRIDGE groups eligible response positions into non-overlapping structural units, including expressions, reasoning steps, and code statements. Given final conditional scores $s_{k,j}=\Delta_{r,k}(j\mid S_{p,k}^{(T)})$, unit utility is computed over core positions:
\begin{equation}
\label{eq:bridge_unit_utility}
q_k(u)=\frac{\sum\nolimits_{j\in u^{\mathrm{core}}}\max(s_{k,j},-\tau)}{(|u^{\mathrm{core}}|+\epsilon_q)^{\alpha}},
\end{equation}
where $\tau$ controls negative-score clipping, $\alpha$ controls length normalization, and $\epsilon_q>0$ ensures numerical stability. Core positions determine utility, while the budget cost $|u|$ counts all unit positions.

Projection first reserves answer positions $\mathcal{F}_k\subseteq\mathcal{R}_k$ within the response budget. With residual capacity $B_{r,k}=b_{r,k}-|\mathcal{F}_k|$, it selects $\mathcal{C}\subseteq\mathcal{U}_k^{\mathrm{free}}$ from unreserved units according to
\begin{equation}
\label{eq:closure_projection}
\max_{\mathcal{C}}\;\sum_{u\in\mathcal{C}}q_k(u)
\quad\text{s.t.}\quad
\sum_{u\in\mathcal{C}}|u|\leq B_{r,k}.
\end{equation}
The resulting response subset satisfies $|S_{r,k}^{\mathrm{cl}}|\leq b_{r,k}$. Projection keeps the selected prompt fixed and uses the precomputed conditional scores. Unit definitions and projection implementation details are given in Appendix~\ref{app:closure_diagnostics}. Algorithm~\ref{alg:BRIDGE} summarizes the complete procedure.

\begin{algorithm}[t]
\caption{BRIDGE}
\label{alg:BRIDGE}
\KwIn{$D,D_{\mathrm{w}},D_{\mathrm{val}},\theta_{\mathrm{old}}$; budgets $\{b_{p,k},b_{r,k}\}_{k=1}^{N}$; $T,\lambda$}
\KwOut{Fine-tuned model $\theta_{\mathrm{new}}$}
$\bar{\theta}\leftarrow\mathrm{Warmup}(\theta_{\mathrm{old}},D_{\mathrm{w}})$\;
$v\leftarrow\mathrm{Direction}(\bar{\theta},\theta_{\mathrm{old}},D_{\mathrm{val}},\lambda)$\;
\For{$k=1,\ldots,N$}{
    $(\widetilde{\beta}_k,\widetilde{\gamma}_k)\leftarrow\mathrm{Score}(\bar{\theta},v,P_k,R_k)$\;
    $S_{r,k}^{(0)}\leftarrow\operatorname{Top}_{b_{r,k}}(\widetilde{\gamma}_k)$\;
    \For{$t=0,\ldots,T-1$}{
        $S_{p,k}^{(t+1)}\leftarrow\operatorname{Top}_{b_{p,k}}\Delta_{p,k}(\cdot\mid S_{r,k}^{(t)})$\;
        $S_{r,k}^{(t+1)}\leftarrow\operatorname{Top}_{b_{r,k}}\Delta_{r,k}(\cdot\mid S_{p,k}^{(t+1)})$\;
    }
    $s_k\leftarrow\Delta_{r,k}(\cdot\mid S_{p,k}^{(T)})$\;
    $S_{r,k}^{\mathrm{cl}}\leftarrow\mathrm{Closure}(R_k,S_{r,k}^{(T)},s_k,b_{r,k})$\;
    $\widetilde{P}_k\leftarrow\mathrm{Compact}(P_k,F_{p,k}\cup S_{p,k}^{(T)})$\;
}
$\widetilde{D}\leftarrow\{(\widetilde{P}_k,R_k,S_{r,k}^{\mathrm{cl}}\cup F_{r,k})\}_{k=1}^{N}$\;
$\theta_{\mathrm{new}}\leftarrow\mathrm{SFT}(\theta_{\mathrm{old}},\widetilde{D})$\;
\end{algorithm}

\subsection{Training Execution}
\label{subsec:training_execution}

After selection, BRIDGE compacts the prompt to $\widetilde{P}_k=\operatorname{Compact}(P_k,F_{p,k}\cup S_{p,k}^{(T)})$ and restores fixed supervised positions through $\widehat{S}_{r,k}=S_{r,k}^{\mathrm{cl}}\cup F_{r,k}$. Prompt compaction shortens the executed sequence from the first Transformer layer. The input $\widetilde{P}_k\Vert R_k$ retains the complete response: only selected positions contribute direct losses, while unsupervised response positions retain gradient paths from later supervised predictions. With $\widetilde{\mathcal{H}}_{k,j}=(\widetilde{P}_k,R_{k,<j})$ and $Z_r=\sum\nolimits_{k=1}^{N}|\widehat{S}_{r,k}|$, final SFT minimizes
\begin{equation}
\label{eq:bridge_training_loss}
\mathcal{L}_{\mathrm{BRIDGE}}=
\frac{1}{Z_r}\sum_{k=1}^{N}
\sum_{j\in\widehat{S}_{r,k}}
\ell(r_{k,j};\theta,\widetilde{\mathcal{H}}_{k,j})
\end{equation}
from $\theta_{\mathrm{old}}$, updating all parameters. Selected prompts and supervision masks remain fixed throughout final SFT.

\textbf{Complexity.} Neural scoring is performed offline, and interaction coefficients are cached across all alternating rounds. For $p=|\mathcal{P}_k|$ and $r=|\mathcal{R}_k|$, interaction assembly requires $O(pr)$ time and storage after neural scoring. Conditional-score aggregation costs $O(T(pb_{r,k}+rb_{p,k}))$, and sorting-based Top-$b$ extraction adds $O(T(p\log p+r\log r))$. The dynamic-programming projection branch costs $O(M_kb_{r,k})$ for $M_k$ structural units. These operations precede final SFT, which executes sequences of length $|F_{p,k}|+b_{p,k}+|R_k|$. Response masking determines which positions contribute direct losses, while forward computation still processes the complete response.

\section{Experiments}
\label{sec:experiments}
Extending the analysis in Section~\ref{sec:coupling_analysis}, we evaluate task adaptation and capability retention under reduced context and supervision budgets. Controlled comparisons isolate coupling benefits; ablations and fidelity analyses assess the interaction surrogate, alternating updates, and structural projection.

\subsection{Experimental Setup}
\label{subsec:experimental_setup}

\textbf{Models and Datasets.} We evaluate BRIDGE on Llama-3.2-3B~\citep{grattafiori2024llama}, Gemma-3-4B-PT~\citep{team2025gemma}, and Qwen3.5-9B-Base~\citep{qwen35blog}. We fine-tune these models on MetaMathQA-40K~\citep{yu2024metamath} for mathematical reasoning, Magicoder~\citep{wei2024magicoder} for code generation, and Alpaca~\citep{taori2023alpaca} for instruction following.

\textbf{Baselines.}  We compare against five groups:
(i) standard and random references (Pretrained, Full SFT, and Random);
(ii) prompt compression (LLMLingua-2~\citep{pan2024llmlingua});
(iii) response-supervision selection
(RHO-1~\citep{lin2024not}, ssToken~\citep{qin2026sstoken},TokenSeek~\citep{zeng2026tokenseek}
and Token Cleaning~\citep{pang2025token});
(iv) two-sided selector combinations, represented by independent
LLM+TC, with additional independent and sequential
compositions; and
(v) forgetting-mitigation methods
(FLOW~\citep{sanyal2025upweighting},
TALR~\citep{lin2026sft},
SAM~\citep{li2024revisiting},
NanoAdam~\citep{zhou2025payattentionsmallweights},
and FAPM~\citep{huang2025mitigating}).
Appendix~\ref{app:baseline_details} describes the implementations,
budget matching, and supplementary mechanism controls.

\textbf{Evaluation.}
Following prior studies, we assess both downstream adaptation and
capability retention~\citep{sanyal2025upweighting,lin2026sft,chen2026algorithm}.
Target-task performance is measured by
MATH accuracy~\citep{hendrycks2021measuringa},
HumanEval pass@1~\citep{chen2021evaluating},
and IFEval strict prompt- and instruction-level
accuracy~\citep{zhou2023instructionfollowing}
for the three domains, respectively.
We assess general-capability retention on
ARC-Challenge~\citep{clark2018think},
HellaSwag~\citep{zellers2019hellaswag},
and MMLU~\citep{hendrycks2021measuring},
supplemented with HumanEval after mathematical fine-tuning
and GSM8K~\citep{cobbe2021training} after code or instruction fine-tuning.
General Avg. averages the four retention benchmarks,
Overall Avg. averages General Avg. and the target-task score.

\textbf{Experimental Configuration.} Optimization hyperparameters are selected separately for each method, base model, and dataset, and are reported in Table~\ref{tab:hyperparams}. Within each backbone and dataset, the selection methods share data splits, AdamW~\citep{loshchilov2019decoupled}, cosine scheduling, and effective batch size $64$. BRIDGE uses three epochs at $2\times10^{-5}$ across all backbones and datasets. BRIDGE uses $75\%$ budgets over eligible positions on each side, the final four layers for gradient scoring, and $T=4$ alternating rounds before structural projection. Two-sided baselines match per-side budgets; one-sided baselines keep the other side intact and match the combined context and supervision count as closely as feasible. Main-table results average seeds $42$, $3407$, and $2027$; supplementary analyses use single runs unless noted. Appendix~\ref{app:implementation} provides implementation and evaluation details.

\subsection{Main Results}
\label{subsec:main_results}

\begin{table*}[t]
\centering\small
\setlength{\tabcolsep}{4.5pt}
\renewcommand{\arraystretch}{1}
\setlength{\aboverulesep}{1.2pt}
\setlength{\belowrulesep}{1.2pt}
\resizebox{\textwidth}{!}{\begin{tabular}{clccccccc}
\toprule
\multicolumn{1}{c}{\multirow{2}{*}{\textbf{Model}}} & \multirow{2}{*}{\textbf{Method}} & \multicolumn{5}{c}{\textbf{General Capability (\%)}} & \multicolumn{1}{c}{\textbf{Target (\%)}} & \multicolumn{1}{c}{\textbf{Overall}} \\
\cmidrule(lr){3-7} \cmidrule(lr){8-8} \cmidrule(lr){9-9}
& & \textbf{ARC-C} & \textbf{HellaSwag} & \textbf{MMLU} & \textbf{HumanEval} & \textbf{Avg.} & \textbf{MATH} & \textbf{Avg.} \\
\midrule
\multirow{10}{*}{\rotatebox{90}{\textbf{Llama-3.2-3B}}} & \color{baselinecolor} Pretrained & \color{baselinecolor} 43.17 & \color{baselinecolor} 54.20 & \color{baselinecolor} 56.64 & \color{baselinecolor} 28.05 & \color{baselinecolor} 45.52 & \color{baselinecolor} 0.62 & \color{baselinecolor} 23.07 \\
& \color{baselinecolor} Full SFT & \color{baselinecolor} $40.73_{\pm 0.36}$ & \color{baselinecolor} $52.18_{\pm 0.17}$ & \color{baselinecolor} $51.84_{\pm 0.29}$ & \color{baselinecolor} $18.29_{\pm 1.06}$ & \color{baselinecolor} $40.76_{\pm 0.24}$ & \color{baselinecolor} $16.66_{\pm 0.52}$ & \color{baselinecolor} $28.71_{\pm 0.29}$ \\
& Random & $41.64_{\pm 0.52}$ & \second{$54.83_{\pm 0.36}$} & $53.80_{\pm 0.26}$ & \second{$30.89_{\pm 0.70}$} & $45.29_{\pm 0.21}$ & $11.27_{\pm 0.87}$ & $28.28_{\pm 0.33}$ \\
& RHO-1\,\venuetag{NeurIPS'24} & $41.27_{\pm 0.27}$ & $53.92_{\pm 0.23}$ & \best{$55.38_{\pm 0.17}$} & $29.07_{\pm 1.27}$ & $44.91_{\pm 0.25}$ & $14.67_{\pm 0.45}$ & $29.79_{\pm 0.35}$ \\
& LLMLingua-2\,\venuetag{ACL'24} & $42.12_{\pm 0.40}$ & $54.56_{\pm 0.19}$ & $54.49_{\pm 0.23}$ & $29.07_{\pm 0.35}$ & $45.06_{\pm 0.26}$ & \second{$14.96_{\pm 0.84}$} & \second{$30.01_{\pm 0.43}$} \\
& ssToken\,\venuetag{ICLR'26} & $40.87_{\pm 0.34}$ & $54.25_{\pm 0.17}$ & $54.69_{\pm 0.34}$ & $29.47_{\pm 0.93}$ & $44.82_{\pm 0.23}$ & $13.89_{\pm 0.65}$ & $29.36_{\pm 0.25}$ \\
& TokenSeek\,\venuetag{ICLR'26} & \second{$42.29_{\pm 0.49}$} & $53.92_{\pm 0.18}$ & $54.07_{\pm 0.35}$ & $27.44_{\pm 0.61}$ & $44.43_{\pm 0.24}$ & $11.91_{\pm 0.77}$ & $28.17_{\pm 0.37}$ \\
& TokenCleaning\,\venuetag{ICML'25} & $42.06_{\pm 0.26}$ & $54.35_{\pm 0.22}$ & $54.64_{\pm 0.35}$ & $29.67_{\pm 0.93}$ & $45.18_{\pm 0.27}$ & $14.17_{\pm 0.57}$ & $29.68_{\pm 0.34}$ \\
& LLM+TC(Ind.) & $41.92_{\pm 0.39}$ & $54.63_{\pm 0.36}$ & $54.56_{\pm 0.29}$ & $30.49_{\pm 1.22}$ & \second{$45.40_{\pm 0.29}$} & $13.10_{\pm 0.40}$ & $29.25_{\pm 0.34}$ \\
\rowcolor{oursrow} \cellcolor{white} & \textbf{BRIDGE (Ours)} & \best{$43.06_{\pm 0.30}$} & \best{$55.95_{\pm 0.25}$} & \second{$55.32_{\pm 0.15}$} & \best{$32.32_{\pm 1.06}$} & \best{$46.66_{\pm 0.20}$} & \best{$15.96_{\pm 0.50}$} & \best{$31.31_{\pm 0.32}$} \\
\midrule
\multirow{10}{*}{\rotatebox{90}{\textbf{Gemma-3-4B-PT}}} & \color{baselinecolor} Pretrained & \color{baselinecolor} 51.37 & \color{baselinecolor} 54.86 & \color{baselinecolor} 59.72 & \color{baselinecolor} 35.37 & \color{baselinecolor} 50.33 & \color{baselinecolor} 4.30 & \color{baselinecolor} 27.32 \\
& \color{baselinecolor} Full SFT & \color{baselinecolor} $40.98_{\pm 0.47}$ & \color{baselinecolor} $51.69_{\pm 0.31}$ & \color{baselinecolor} $54.74_{\pm 0.21}$ & \color{baselinecolor} $30.49_{\pm 0.61}$ & \color{baselinecolor} $44.48_{\pm 0.24}$ & \color{baselinecolor} $22.92_{\pm 0.74}$ & \color{baselinecolor} $33.70_{\pm 0.38}$ \\
& Random & $43.03_{\pm 0.38}$ & $53.37_{\pm 0.32}$ & $54.47_{\pm 0.26}$ & $32.11_{\pm 1.27}$ & $45.75_{\pm 0.27}$ & $18.03_{\pm 0.41}$ & $31.89_{\pm 0.33}$ \\
& RHO-1\,\venuetag{NeurIPS'24} & \second{$52.96_{\pm 0.34}$} & $54.68_{\pm 0.27}$ & $56.01_{\pm 0.18}$ & $36.59_{\pm 1.22}$ & $50.06_{\pm 0.24}$ & $22.97_{\pm 0.80}$ & $36.51_{\pm 0.37}$ \\
& LLMLingua-2\,\venuetag{ACL'24} & $52.02_{\pm 0.30}$ & \second{$55.93_{\pm 0.25}$} & $56.14_{\pm 0.15}$ & $36.79_{\pm 0.70}$ & \second{$50.22_{\pm 0.21}$} & $22.83_{\pm 0.66}$ & \second{$36.53_{\pm 0.31}$} \\
& ssToken\,\venuetag{ICLR'26} & $51.82_{\pm 0.44}$ & $54.64_{\pm 0.25}$ & $56.28_{\pm 0.19}$ & $35.98_{\pm 0.61}$ & $49.68_{\pm 0.30}$ & \second{$23.27_{\pm 0.61}$} & $36.48_{\pm 0.34}$ \\
& TokenSeek\,\venuetag{ICLR'26} & $50.54_{\pm 0.30}$ & $55.32_{\pm 0.28}$ & $57.12_{\pm 0.24}$ & $33.74_{\pm 1.27}$ & $49.18_{\pm 0.25}$ & $22.47_{\pm 0.88}$ & $35.82_{\pm 0.31}$ \\
& TokenCleaning\,\venuetag{ICML'25} & $51.56_{\pm 0.36}$ & $54.91_{\pm 0.17}$ & $55.41_{\pm 0.22}$ & \second{$37.80_{\pm 1.06}$} & $49.92_{\pm 0.18}$ & $21.73_{\pm 0.83}$ & $35.83_{\pm 0.33}$ \\
& LLM+TC(Ind.) & $50.23_{\pm 0.55}$ & $50.17_{\pm 0.27}$ & \second{$58.03_{\pm 0.24}$} & $35.98_{\pm 1.06}$ & $48.60_{\pm 0.24}$ & $19.26_{\pm 0.72}$ & $33.93_{\pm 0.28}$ \\
\rowcolor{oursrow} \cellcolor{white} & \textbf{BRIDGE (Ours)} & \best{$53.53_{\pm 0.26}$} & \best{$57.17_{\pm 0.33}$} & \best{$58.16_{\pm 0.16}$} & \best{$39.43_{\pm 0.70}$} & \best{$52.07_{\pm 0.26}$} & \best{$24.08_{\pm 0.46}$} & \best{$38.08_{\pm 0.33}$} \\
\midrule
\multirow{10}{*}{\rotatebox{90}{\textbf{Qwen3.5-9B}}} & \color{baselinecolor} Pretrained & \color{baselinecolor} 54.27 & \color{baselinecolor} 54.26 & \color{baselinecolor} 72.01 & \color{baselinecolor} 61.59 & \color{baselinecolor} 60.53 & \color{baselinecolor} 40.06 & \color{baselinecolor} 50.30 \\
& \color{baselinecolor} Full SFT & \color{baselinecolor} $53.21_{\pm 0.32}$ & \color{baselinecolor} $57.83_{\pm 0.24}$ & \color{baselinecolor} $71.09_{\pm 0.34}$ & \color{baselinecolor} $59.15_{\pm 1.22}$ & \color{baselinecolor} $60.32_{\pm 0.28}$ & \color{baselinecolor} $49.44_{\pm 0.43}$ & \color{baselinecolor} $54.88_{\pm 0.27}$ \\
& Random & $51.82_{\pm 0.43}$ & $57.17_{\pm 0.16}$ & $71.72_{\pm 0.21}$ & $54.88_{\pm 0.61}$ & $58.90_{\pm 0.24}$ & $45.83_{\pm 0.85}$ & $52.36_{\pm 0.31}$ \\
& RHO-1\,\venuetag{NeurIPS'24} & $50.88_{\pm 0.27}$ & $58.03_{\pm 0.14}$ & \best{$73.77_{\pm 0.18}$} & $55.69_{\pm 0.35}$ & $59.59_{\pm 0.24}$ & $47.32_{\pm 0.50}$ & $53.46_{\pm 0.32}$ \\
& LLMLingua-2\,\venuetag{ACL'24} & \second{$53.47_{\pm 0.34}$} & \second{$59.03_{\pm 0.33}$} & $71.97_{\pm 0.27}$ & \best{$56.30_{\pm 0.70}$} & \second{$60.19_{\pm 0.18}$} & \second{$49.63_{\pm 0.64}$} & \second{$54.91_{\pm 0.24}$} \\
& ssToken\,\venuetag{ICLR'26} & $53.04_{\pm 0.25}$ & $58.17_{\pm 0.15}$ & $71.43_{\pm 0.22}$ & $52.44_{\pm 0.61}$ & $58.77_{\pm 0.25}$ & $49.24_{\pm 0.65}$ & $54.01_{\pm 0.36}$ \\
& TokenSeek\,\venuetag{ICLR'26} & $50.63_{\pm 0.50}$ & $57.71_{\pm 0.22}$ & $71.38_{\pm 0.25}$ & $51.22_{\pm 0.61}$ & $57.73_{\pm 0.24}$ & $46.72_{\pm 0.60}$ & $52.23_{\pm 0.29}$ \\
& TokenCleaning\,\venuetag{ICML'25} & $52.42_{\pm 0.40}$ & $58.87_{\pm 0.31}$ & $72.27_{\pm 0.16}$ & $51.83_{\pm 1.06}$ & $58.85_{\pm 0.28}$ & $48.12_{\pm 0.63}$ & $53.48_{\pm 0.32}$ \\
& LLM+TC(Ind.) & $52.76_{\pm 0.30}$ & $58.43_{\pm 0.25}$ & \second{$73.57_{\pm 0.19}$} & $53.25_{\pm 0.93}$ & $59.50_{\pm 0.20}$ & $48.46_{\pm 0.48}$ & $53.98_{\pm 0.34}$ \\
\rowcolor{oursrow} \cellcolor{white} & \textbf{BRIDGE (Ours)} & \best{$54.78_{\pm 0.37}$} & \best{$59.34_{\pm 0.21}$} & $73.38_{\pm 0.22}$ & \second{$56.10_{\pm 0.61}$} & \best{$60.90_{\pm 0.28}$} & \best{$50.40_{\pm 0.61}$} & \best{$55.65_{\pm 0.41}$} \\
\bottomrule
\end{tabular}}
\vspace{-0.1in}
\caption{Results of token selection methods on downstream benchmarks, reported in accuracy (\%). Excluding \textit{Pretrained} and \textit{Full SFT}, the best and second-best results in each column are highlighted in \textcolor{resultbest}{\textbf{red}} and \textcolor{resultsecond}{\underline{blue}}, respectively.}
\label{tab:main_results}
\vspace{-0.1in}
\end{table*}

\textbf{Joint gains across backbones and domains.}
BRIDGE leads evaluated selection methods in MATH accuracy
and General Avg. on all three backbones
(Table~\ref{tab:main_results}), improving Overall Avg. over the
strongest selection baseline by $1.30$, $1.55$, and $0.74$ percentage
points on Llama, Gemma, and Qwen, respectively.
It spans all nine backbone--task combinations, including code generation
and instruction following
(Appendices~\ref{app:code_generation}
and~\ref{app:instruction_following}).

\textbf{Benefits of coordinated selection.}
BRIDGE improves both MATH accuracy and General Avg. over independent
LLM+TC across backbones. It also outperforms the additional independent
and sequential combinations evaluated on Llama
(Appendix~\ref{app:baseline_combinations}), supporting the value of
coordinated prompt--response selection.

\textbf{Competitive adaptation--retention trade-offs.}
In mathematical fine-tuning, BRIDGE improves Overall Avg. over
Full SFT by $0.77$--$4.38$ points at $75\%$ per-side budgets and
exceeds pretrained General Avg. on all three backbones.
On Llama, it trades slightly lower MATH accuracy for stronger retention.
BRIDGE also achieves retention comparable to FLOW with higher target
accuracy and Overall Avg., although the comparison is not
compute-matched (Appendix~\ref{app:forgetting_baselines}).

\subsection{Ablation and Sensitivity Analysis}
\label{subsec:design_analysis}

\textbf{Component contributions.} Single-run ablations on Llama-3.2-3B (Table~\ref{tab:design_analysis}) show Overall Avg. drops of $1.45$, $0.75$, $0.71$, and $0.57$ points without scoring warmup, the L2 anchor, response-history utility, and structural projection, respectively. With the scoring checkpoint fixed, pretrained initialization outperforms warmup initialization by $0.74$ points, supporting warmup as a scoring probe.

\begin{figure*}[t]
\centering
\includegraphics[width=\linewidth]{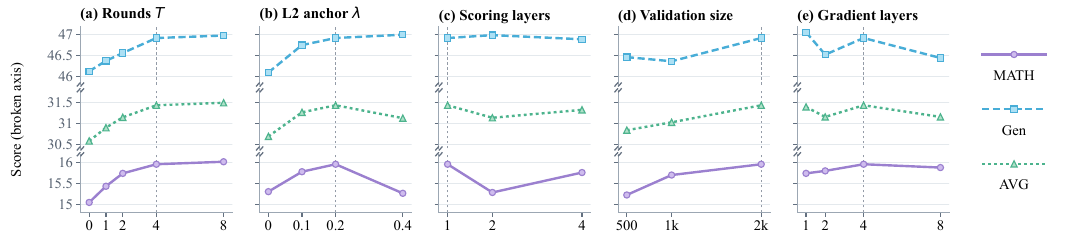}
\vspace{-0.3in}
\caption{Single-run parameter sensitivity on Llama-3.2-3B.}
\label{fig:design_sweeps}
\vspace{-0.15in}
\end{figure*}

\textbf{Alternating selection.} With scoring and budgets fixed, Overall Avg. increases from $30.58$ for Independent and $30.90$ for One-Pass to $31.44$ at $T=4$; $T=8$ adds only $0.06$ point, supporting the four-round default (Figure~\ref{fig:design_sweeps}). Post-Closure refresh leaves Overall Avg. unchanged at reported precision. Appendices~\ref{app:alternating_rounds} and~\ref{app:closure_diagnostics} detail round sensitivity and projection diagnostics.

\textbf{Scoring sensitivity.} The default configuration---$\lambda=0.2$, one attention layer, four gradient-scoring layers, and 1,963 validation examples---achieves the highest MATH score and Overall Avg.\ in the Llama hyperparameter sweeps (Appendix~\ref{app:scoring_hyperparameters}). Appendices~\ref{app:l2_anchor_ablation} and~\ref{app:scoring_anchor} detail anchor sensitivity and scoring-checkpoint comparisons.

\begin{table}[t]
\centering\small
\setlength{\tabcolsep}{4pt}
\resizebox{\columnwidth}{!}{\begin{tabular}{lccc}
\toprule
\textbf{Variant} & \textbf{MATH} & \textbf{General Avg.} & \textbf{Overall Avg.} \\
\midrule
BRIDGE & 15.96 & 46.92 & 31.44 \\
\quad w/o warmup for scoring & 13.96 & 46.02 & 29.99 \\
\quad SFT initialized from warmup & 15.42 & 45.98 & 30.70 \\
\quad w/o L2 anchor & 15.30 & 46.09 & 30.69 \\
\quad w/o response-history term & 15.22 & 46.24 & 30.73 \\
\quad w/o structure projection & 15.20 & 46.53 & 30.87 \\
\quad Post-Closure refresh & 15.93 & 46.94 & 31.44 \\
\bottomrule
\end{tabular}}
\vspace{-0.1in}
\caption{Single-run design comparisons on Llama-3.2-3B under fixed token budgets.}
\label{tab:design_analysis}
\vspace{-0.1in}
\end{table}

\subsection{Selection-Level Fidelity of the Surrogate}
\label{subsec:selection_fidelity}

We evaluate how accurately the conditional surrogate scores
$\Delta_p$ and $\Delta_r$ preserve reference rankings, selected subsets,
and utilities. We use 200 randomly sampled MetaMathQA examples at the
Llama-3.2-3B warmup checkpoint with $\rho_p=\rho_r=0.75$.
The reference scores are
\begin{equation}
\begin{aligned}
\Delta^{\mathrm{exact}}_{p,k}(i\mid S_r)
&= \sum\nolimits_{j\in S_r}\beta_{k,i,j},\\
\Delta^{\mathrm{exact}}_{r,k}(j\mid S_p)
&= v^\top g_{k,j}(\widetilde{P}_k).
\end{aligned}
\end{equation}
Here, the prompt reference aggregates the length-preserving perturbation
interactions defined in Eq.~\eqref{eq:exact_interaction_reference},
while the response reference directly recomputes gradient alignment
under the compacted prompt $\widetilde{P}_k$.
We compare Attention, which omits the validation-directed multiplier
$a_{k,j}$, with Attention $\times a$, which uses the full surrogate
in Eq.~\eqref{eq:conditional_scores}.
We report Spearman correlation, Top-$b$ overlap, Jaccard similarity,
and normalized regret relative to the reference Top-$b$ selection.

\begin{table}[t]
\centering\small
\setlength{\tabcolsep}{2.2pt}
\resizebox{\columnwidth}{!}{%
\begin{tabular}{llcccc}
\toprule
\textbf{Side}
& \textbf{Proxy}
& \makecell[c]{\textbf{Spearman}\\$\uparrow$}
& \makecell[c]{\textbf{Top-$b$}\\\textbf{Overlap} $\uparrow$}
& \makecell[c]{\textbf{Jaccard}\\$\uparrow$}
& \makecell[c]{\textbf{Norm.}\\\textbf{Regret} $\downarrow$} \\
\midrule
\multirow{2}{*}{Prompt}
& Attention
& $-0.0528$ & $0.7094$ & $0.5509$ & $0.5905$ \\
& Attention $\times a$
& $\mathbf{0.9123}$ & $\mathbf{0.9462}$
& $\mathbf{0.9192}$ & $\mathbf{0.0347}$ \\
\midrule
\multirow{2}{*}{Response}
& Attention
& $-0.0409$ & $0.7639$ & $0.6187$ & $0.5182$ \\
& Attention $\times a$
& $\mathbf{0.9463}$ & $\mathbf{0.9667}$
& $\mathbf{0.9362}$ & $\mathbf{0.0251}$ \\
\bottomrule
\end{tabular}%
}
\vspace{-0.1in}
\caption{Selection-level fidelity to reference scores.}
\label{tab:selection_fidelity}
\vspace{-0.2in}
\end{table}

Attention alone yields near-zero rank correlations.
Including the validation-directed multiplier raises Spearman correlation
above $0.91$ and Top-$b$ overlap above $94\%$, reducing normalized
regret below $0.035$ on both sides
(Table~\ref{tab:selection_fidelity}).
The full surrogate therefore closely preserves the reference rankings
and selected subsets with low reference-utility loss.
Appendix~\ref{app:state_rescoring} further examines this agreement
across alternating rounds by recomputing response scores under
physically compacted prompts.

\subsection{Structure-Aware Projection}
\label{subsec:structure_projection}

Structure-aware projection reallocates response supervision across
local structural units under the same response-budget cap.
It uses existing token scores and requires no additional model inference
or gradient computation.
We compare Score-only and Score+Closure on Llama-3.2-3B,
holding prompt selection and final training fixed.

Score+Closure improves MATH accuracy, General Avg., and Overall Avg.
by $0.76$, $0.39$, and $0.57$ percentage points, respectively
(Table~\ref{tab:design_analysis}).
It achieves these gains while retaining $83\%$ of the pre-projection surrogate utility and using $97\%$ of the response budget, supporting structural projection beyond token-level score optimization.
Appendix~\ref{app:closure_diagnostics} provides projection details,
coverage diagnostics, and a case study
(Figure~\ref{fig:structure_projection}).

\subsection{Computational Cost Analysis}
\label{subsec:efficiency}

\begin{figure}[t]
\vspace{-0.25in}
\centering
\includegraphics[width=\columnwidth]{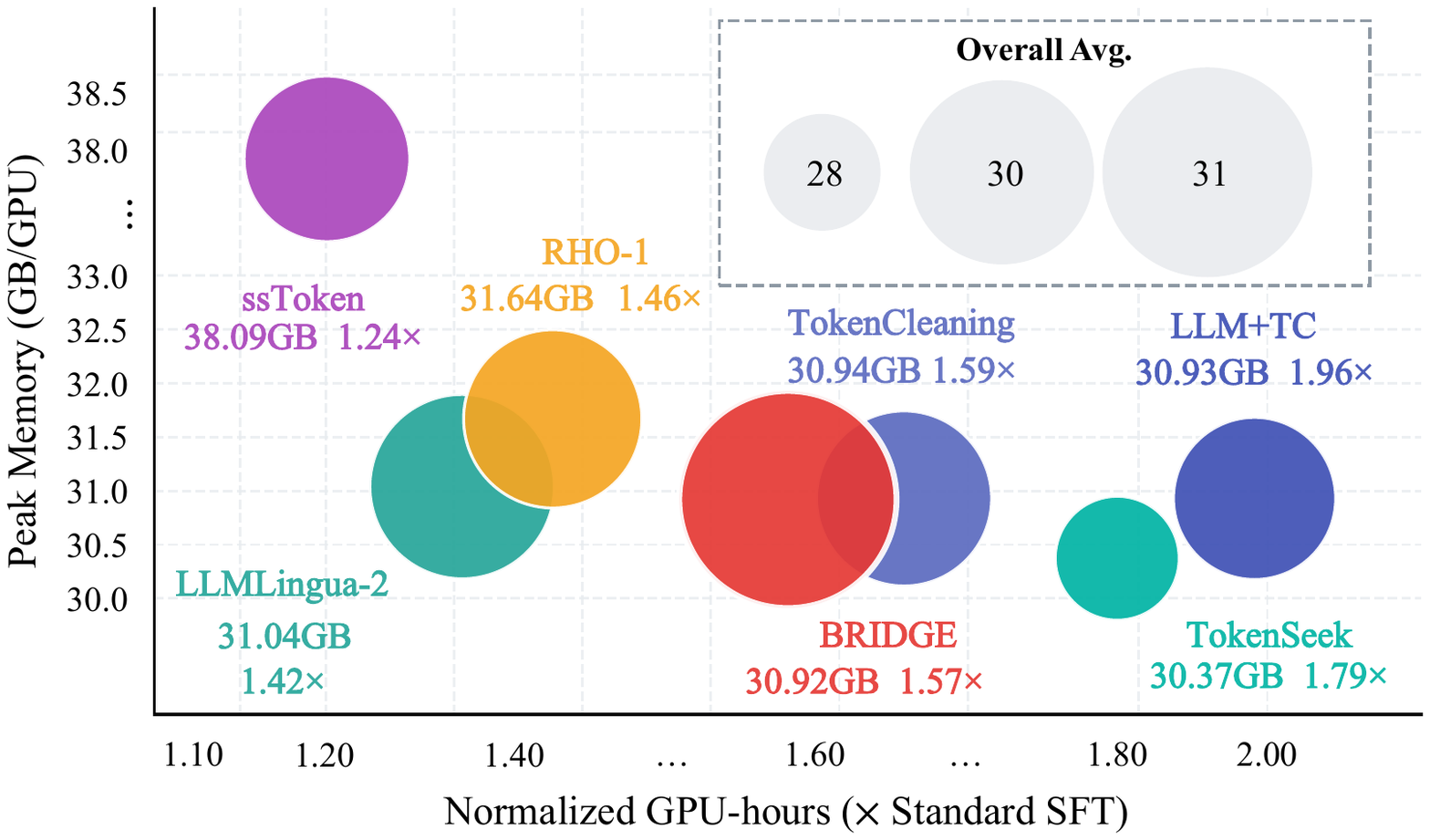}
\vspace{-0.5in}
\caption{Efficiency and performance comparison.}
\label{fig:efficiency}
\vspace{-0.2in}
\end{figure}

Figure~\ref{fig:efficiency} compares cost and peak memory on Llama-3.2-3B under matched settings. Costs are reported in A100-80GB-equivalent GPU-hours; Appendix~\ref{app:efficiency_details} details timing and stage-wise throughput. BRIDGE reduces final-SFT cost from $2.29$ to $2.14$ GPU-hours ($6.6\%$) and peak memory from $32.67$ to $30.92$ GB ($5.4\%$). Prompt compaction shortens execution, while unsupervised response positions remain in the causal history. Warmup and scoring/selection add $1.45$ GPU-hours, bringing end-to-end cost to $3.59$ GPU-hours; preprocessing exceeds final-SFT savings over three epochs.

\section{Conclusion}
Token selection is a form of learning-signal design: retained context and supervised targets jointly determine how an example contributes to learning. BRIDGE establishes the value of keeping token valuation consistent with this retained configuration, with larger coordination gains under stronger compression. Our analysis further identifies selection-level fidelity as a practical criterion for assessing interaction surrogates. Together with the joint gains in task adaptation and average capability retention, these findings position coordinated selection as a means of shaping the adaptation--retention balance through the training data itself. Future work can extend this principle to selection that adapts to evolving model states during fine-tuning, using selective interaction refresh to control scoring costs.

\section*{Limitations}

Our study has several limitations.
First, our experiments span mathematical reasoning, code generation, and instruction following with 3B--9B backbones, while detailed coupling analyses are limited to mathematical reasoning. Extending these analyses to other domains and larger models would provide a broader assessment of the benefits of coordinated selection.
Second, the interaction surrogate preserves conditional rankings and Top-$b$ subsets with high fidelity on 200 MetaMathQA examples at the Llama-3.2-3B scoring checkpoint used in our evaluation. Additional selection-level diagnostics across models, tasks, and scoring checkpoints would provide a more comprehensive assessment of surrogate fidelity.
Third, our experiments apply fixed prompt and response retention ratios across examples. Difficulty-aware budget allocation could extend the current selection framework to accommodate differences in example difficulty.
Finally, BRIDGE reduces final-SFT cost in our three-epoch cost evaluation. The one-time preprocessing overhead from warmup, scoring, and selection increases end-to-end cost. Longer training or reuse of the selected subsets under compatible scoring settings may amortize this overhead more effectively. Reducing preprocessing cost and dependence on held-out validation data would improve applicability in resource-constrained settings.




\bibliography{custom}

\clearpage
\appendix

The appendix follows the theoretical and empirical development of the main
paper. Appendix~\ref{app:theoretical_details} provides supplementary proofs;
Appendix~\ref{app:implementation} summarizes the shared experimental protocol;
Appendix~\ref{app:selection_conditioned_utility} supplements the evidence for
selection-conditioned utility; Appendix~\ref{app:additional_tasks} reports
additional results and capability retention; Appendix~\ref{app:design_ablations}
provides ablation and sensitivity analyses; Appendix~\ref{app:interaction_approximation}
examines interaction-surrogate fidelity; Appendix~\ref{app:structural_projection}
details structural projection and supervision coverage; and
Appendix~\ref{app:efficiency_details} gives the computational cost breakdown.

\section{Theoretical Details}
\label{app:theoretical_details}

We analyze the local learning interpretation of directional scoring,
selection-state dependence, stability under scoring error, and the
monotonicity of alternating selection. These results concern the
token-level stage. We separately characterize prompt stability after
structural projection.

\subsection{Local Interpretation of Directional Scoring}
\label{app:local_descent}

Let \(\phi=\theta_{\mathcal G}\) and
\(\bar\phi=\bar\theta_{\mathcal G}\), with parameters outside
\(\mathcal G\) fixed at their warmup values. Fix the normalization
constants at the scoring checkpoint:
\[
c_t=\frac{1-\lambda}{\|g_{\mathrm{target}}\|_2+\epsilon},
\qquad
c_a=\frac{\lambda}{\|g_{\mathrm{anchor}}\|_2+\epsilon}.
\]
Writing \(L_{\mathrm{val}}^{\mathcal G}(\phi)\) for the restricted
validation loss, define
\begin{equation}
\label{eq:local_scoring_criterion}
\Psi(\phi)=c_tL_{\mathrm{val}}^{\mathcal G}(\phi)
+\frac{c_a}{2}\|\phi-\theta_{\mathrm{old},\mathcal G}\|_2^2.
\end{equation}
Then
\[
\nabla\Psi(\bar\phi)
=c_tg_{\mathrm{target}}+c_ag_{\mathrm{anchor}}=v.
\]
Thus, the scoring direction is the gradient of a local
validation--drift criterion.

For a pre-projection selection \(\mathcal S\), let
\(W_{r,k}=S_{r,k}\cup F_{r,k}\) and
\(Z(\mathcal S)=\sum_k|W_{r,k}|>0\). Its training-loss gradient in
the scoring subspace is
\begin{equation}
\label{eq:local_selected_gradient}
G_{\mathcal S}=
\frac{1}{Z(\mathcal S)}
\sum_{k=1}^{N}\sum_{j\in W_{r,k}}
g_{k,j}(\widetilde P_k).
\end{equation}
Let \(U_{\mathrm{loc}}(\mathcal S)\) and
\(U_{\mathrm{fixed}}(\mathcal S)\) denote the directional contributions
from selected eligible positions and fixed positions, respectively,
with
\(U_{\mathrm{fixed}}(\mathcal S)
=\sum_k\sum_{j\in F_{r,k}}v^\top g_{k,j}(\widetilde P_k)\).
Then
\begin{equation}
\label{eq:local_full_alignment}
v^\top G_{\mathcal S}
=\frac{U_{\mathrm{loc}}(\mathcal S)
+U_{\mathrm{fixed}}(\mathcal S)}{Z(\mathcal S)}.
\end{equation}
Fixed eligible-token budgets keep \(Z(\mathcal S)\) constant,
while fixed-position gradients remain dependent on the compacted prompt.
Common positive learning-rate and loss-normalization factors are omitted
from the local utility in Eq.~\eqref{eq:local_utility} under these fixed
budgets.

For a local update \(\phi^+=\bar\phi-\eta G_{\mathcal S}\), assume
that \(\Psi\) has an \(L\)-Lipschitz gradient, with \(L>0\), on a
neighborhood containing the update segment. Smoothness gives
\begin{equation}
\label{eq:local_descent_bound}
\Psi(\phi^+)-\Psi(\bar\phi)
\le -\eta v^\top G_{\mathcal S}
+\frac{L\eta^2}{2}\|G_{\mathcal S}\|_2^2.
\end{equation}
If \(v^\top G_{\mathcal S}>0\), strict descent follows whenever
\[
0<\eta<
\frac{2v^\top G_{\mathcal S}}{L\|G_{\mathcal S}\|_2^2}.
\]

The token score has the same local interpretation. Let
\(\ell_{k,j}(\phi;Q)\) be the token negative log-likelihood under
prompt \(Q\) and the complete response prefix, with the same parameter
restriction. The chain rule gives
\begin{equation}
\label{eq:token_directional_identity}
\begin{aligned}
a_{k,j}&=v^\top g_{k,j}(P_k)\\
&=\left.\frac{d}{dt}\ell_{k,j}(\bar\phi+tv;P_k)\right|_{t=0}.
\end{aligned}
\end{equation}
A token-gradient update changes \(\Psi\) by
\(-\eta a_{k,j}\) to first order, so positive scores favor local
descent. This interpretation applies at the warmup checkpoint within
the scoring subspace; final SFT starts from
\(\theta_{\mathrm{old}}\).

\subsection{Selection-State Dependence}
\label{app:selection_state_theory}
\label{app:actual_state_boundary}
\label{app:recovered_utility}

Fix one example. Let \(\mathcal X\) contain the eligible candidates
on the selected side and \(\mathcal Y\) those on the opposite side.
For a retained state \(V\subseteq\mathcal Y\), define
\begin{equation}
\label{eq:side_agnostic_score}
s_x(V)
=
a_x+\sum_{y\in V}B_{xy},
\qquad x\in\mathcal X.
\end{equation}
This includes both BRIDGE conditional scores. For prompt selection,
\begin{equation}
\begin{aligned}
\mathcal X &= \mathcal P_k, & \mathcal Y &= \mathcal R_k, \\
a_x &= 0, & B_{xy} &= \widetilde{\beta}_{k,x,y}.
\end{aligned}
\end{equation}
For response selection,
\begin{equation}
\begin{aligned}
\mathcal X &= \mathcal R_k, & \mathcal Y &= \mathcal P_k, \\
a_x &= \widetilde{\gamma}_{k,x}, & B_{xy} &= \widetilde{\beta}_{k,y,x}.
\end{aligned}
\end{equation}

For a size-\(b\) subset \(S\subseteq\mathcal X\), let
\begin{equation}
\label{eq:subset_utility}
U_V(S)=\sum_{x\in S}s_x(V).
\end{equation}
Define the full-state and conditional selections as
\begin{equation}
\label{eq:full_state_selection}
\begin{aligned}
S^F&=\operatorname{Top}_b
\{s_x(\mathcal Y):x\in\mathcal X\},\\
S^C(V)&=\operatorname{Top}_b
\{s_x(V):x\in\mathcal X\}.
\end{aligned}
\end{equation}
For \(1\le b<|\mathcal X|\), the full-state selection has margin
\begin{equation}
\label{eq:actual_margin}
\lambda(V)=
\min_{\substack{x\in S^F\\x'\in\mathcal X\setminus S^F}}
[s_x(V)-s_{x'}(V)].
\end{equation}

\begin{lemma}[Selection-state boundary and recovered utility]
\label{lem:actual_state_boundary}
The subset \(S^F\) maximizes \(U_V\) over size-\(b\) subsets if and
only if \(\lambda(V)\ge0\). Conditional reselection recovers
\begin{equation}
\label{eq:recovered_utility}
R(V)=U_V(S^C(V))-U_V(S^F)\ge0,
\end{equation}
with \(R(V)>0\) whenever \(\lambda(V)<0\).
\end{lemma}

\begin{proof}
If \(\lambda(V)\ge0\), every selected candidate scores at least as
highly as every unselected candidate, making \(S^F\) optimal.
If \(\lambda(V)<0\), some \(x\in S^F\) and \(x'\notin S^F\)
satisfy \(s_{x'}(V)>s_x(V)\). Exchanging them increases utility by
\(s_{x'}(V)-s_x(V)>0\).
The conditional Top-\(b\) subset maximizes \(U_V\), giving both
claims about \(R(V)\).
\end{proof}

At \(b=0\) or \(b=|\mathcal X|\), the unique feasible subset is
optimal. For intermediate budgets, \(\lambda(V)\) identifies when
a changed opposite-side state preserves the original selection.

\paragraph{Budget robustness and recovered utility.}
\label{app:budget_robustness}
For \(n=|\mathcal Y|\), sort
\(d_{xx',y}=B_{xy}-B_{x'y}\) increasingly for each
selected--unselected pair. Define
\begin{equation}
\label{eq:kappa_q}
\kappa_q=
\min_{\substack{x\in S^F\\x'\in\mathcal X\setminus S^F}}
\left[a_x-a_{x'}+\sum_{\ell=1}^{q}d_{xx',(\ell)}\right].
\end{equation}
For each pair, choosing the \(q\) smallest differences minimizes its
margin over size-\(q\) states. Interchanging the two finite minima gives
\begin{equation}
\label{eq:worst_state_margin}
\min_{\substack{V\subseteq\mathcal Y\\|V|=q}}\lambda(V)=\kappa_q.
\end{equation}
Together with Lemma~\ref{lem:actual_state_boundary}, this proves
Theorem~\ref{thm:selection_state_robustness}.
For prompts, \(a_x=0\), and \(\kappa_q/q\) is the minimum over
pairs of the averages of their \(q\) smallest differences.
Each average is nondecreasing in \(q\), as is their pointwise
minimum, proving the normalized robustness claim in
Section~\ref{sec:coupling_analysis}.

To express recovered utility, sort scores inside \(S^F\) increasingly
as \(c_1\le\cdots\le c_b\), and scores outside it decreasingly as
\(o_1\ge\cdots\ge o_{|\mathcal X|-b}\). With
\(r=\min\{b,|\mathcal X|-b\}\),
\begin{equation}
\label{eq:recovered_utility_swap}
R(V)=\sum_{\ell=1}^{r}\max\{0,o_\ell-c_\ell\}.
\end{equation}
For any fixed number of exchanges, the best gain replaces the
lowest-scoring selected candidates with the highest-scoring unselected
candidates. Since \(o_\ell-c_\ell\) is nonincreasing, the optimum
includes exactly the positive differences, with zero differences
optional.

\subsection{Selection Guarantees under Scoring Error}
\label{app:topk_stability}
\label{app:reference_utility_condition}

At a fixed retained state \(V\), let \(\widehat s_x(V)=s_x(V)\)
be the BRIDGE surrogate score and \(s_x^{\mathrm{ref}}(V)\) an
independently computed reference score. Assume
\begin{equation}
\label{eq:reference_error}
\max_{x\in\mathcal X}
|\widehat s_x(V)-s_x^{\mathrm{ref}}(V)|
\le\varepsilon.
\end{equation}
For \(1\le b<|\mathcal X|\), define
\[
\begin{aligned}
\widehat S^C&=\operatorname{Top}_b
\{\widehat s_x(V):x\in\mathcal X\},\\
S^\star&=\operatorname{Top}_b
\{s_x^{\mathrm{ref}}(V):x\in\mathcal X\},
\end{aligned}
\]
using a common deterministic tie-breaking rule.
Let \(\widehat U_V\) and \(U_V^{\mathrm{ref}}\) be the corresponding
subset utilities. The reference boundary margin is
\begin{equation}
\label{eq:topk_reference_margin}
\mu=
\min_{\substack{x\in S^\star\\x'\in\mathcal X\setminus S^\star}}
[s_x^{\mathrm{ref}}(V)-s_{x'}^{\mathrm{ref}}(V)].
\end{equation}
Relative to the full-state subset \(S^F\), define
\[
\begin{aligned}
\widehat R(V)
&=\widehat U_V(\widehat S^C)-\widehat U_V(S^F),\\
\Delta_{\mathrm{ref}}(V)
&=U_V^{\mathrm{ref}}(\widehat S^C)-U_V^{\mathrm{ref}}(S^F).
\end{aligned}
\]

\begin{proposition}[Selection stability and reference utility]
\label{prop:topk_stability}
\label{prop:reference_utility_improvement}
\label{prop:reference_selection_regret}
Under Eq.~\eqref{eq:reference_error}, if \(\mu>2\varepsilon\),
then \(\widehat S^C=S^\star\).
For \(m=|\widehat S^C\setminus S^F|\),
\begin{equation}
\label{eq:reference_gain_bound}
\Delta_{\mathrm{ref}}(V)
\ge\widehat R(V)-2m\varepsilon.
\end{equation}
Thus, \(\widehat R(V)>2m\varepsilon\) guarantees positive reference
improvement. With \(m_\star=|S^\star\setminus\widehat S^C|\),
the reference selection regret satisfies
\begin{equation}
\label{eq:reference_selection_regret_bound}
0\le U_V^{\mathrm{ref}}(S^\star)
-U_V^{\mathrm{ref}}(\widehat S^C)
\le2m_\star\varepsilon.
\end{equation}
\end{proposition}

\begin{proof}
For every reference-selected candidate \(x\) and unselected
candidate \(x'\),
\[
\begin{aligned}
\widehat s_x(V)-\widehat s_{x'}(V)
&\ge s_x^{\mathrm{ref}}(V)-s_{x'}^{\mathrm{ref}}(V)
-2\varepsilon\\
&\ge\mu-2\varepsilon.
\end{aligned}
\]
A positive boundary therefore preserves the reference Top-\(b\) set.

For the gain bound, let \(e_x=s_x^{\mathrm{ref}}(V)-\widehat s_x(V)\).
Shared candidates cancel, so
\[
\begin{aligned}
\Delta_{\mathrm{ref}}(V)-\widehat R(V)
&=\sum_{x\in\widehat S^C\setminus S^F}e_x
-\sum_{x\in S^F\setminus\widehat S^C}e_x\\
&\ge-2m\varepsilon.
\end{aligned}
\]
For regret, reference optimality gives the lower bound, while surrogate
optimality gives
\(\widehat U_V(S^\star)-\widehat U_V(\widehat S^C)\le0\).
The two differing sets each contain \(m_\star\) candidates, whose
total error contribution is at most \(2m_\star\varepsilon\).
\end{proof}

These results relate score error to the selection boundary, recovered
utility, and the number of exchanged candidates.
Appendix~\ref{app:state_rescoring} evaluates state-conditioned reference
scores after physical prompt compaction.

\subsection{Monotonicity of Alternating Selection}
\label{app:alternating_monotonicity}

The fixed surrogate objective in
Eq.~\eqref{eq:surrogate_selection_objective} is
\begin{equation}
U_{\mathrm{sur}}(S_p,S_r)
=
\sum_{k=1}^{N}\sum_{j\in S_{r,k}}
\left[
\widetilde{\gamma}_{k,j}
+\sum_{i\in S_{p,k}}\widetilde{\beta}_{k,i,j}
\right].
\end{equation}
Alternating selection holds \(\widetilde{\beta}\) and
\(\widetilde{\gamma}\) fixed and maximizes the objective over one
subset at a time.

\begin{proposition}[Monotonicity of alternating selection]
\label{prop:alternating_monotonicity}
Each update in Eq.~\eqref{eq:alternating_updates} preserves or
increases \(U_{\mathrm{sur}}\). For \(t\ge1\),
\begin{equation}
U_{\mathrm{sur}}(S_p^{(t+1)},S_r^{(t)})
\ge U_{\mathrm{sur}}(S_p^{(t)},S_r^{(t)}),
\end{equation}
and, for \(t\ge0\),
\begin{equation}
U_{\mathrm{sur}}(S_p^{(t+1)},S_r^{(t+1)})
\ge U_{\mathrm{sur}}(S_p^{(t+1)},S_r^{(t)}).
\end{equation}
The first prompt update maximizes the objective conditional on the
initialized response subset \(S_r^{(0)}\).
\end{proposition}

\begin{proof}
With the response subset fixed, the prompt-dependent terms are
\[
\sum_{k=1}^{N}\sum_{i\in S_{p,k}}
\Delta_{p,k}(i\mid S_{r,k}^{(t)}).
\]
The prompt update exactly maximizes these terms by selecting the
\(b_{p,k}\) largest scores. The response update similarly maximizes
\[
\sum_{k=1}^{N}\sum_{j\in S_{r,k}}
\Delta_{r,k}(j\mid S_{p,k}^{(t+1)})
\]
by selecting the \(b_{r,k}\) largest scores.
\end{proof}

The result applies to the fixed surrogate used for coordinate updates.
The ideal bilevel formulation in Eq.~\eqref{eq:ideal_selection}
involves hypergradient computation~\citep{grazzi2020hypergradient}.
Because feasible selections are finite, continued alternating updates
eventually stabilize the objective value. Stabilization of the subsets
additionally requires a tie-breaking rule that excludes equal-value
cycles.

\subsection{Prompt Stability after Structural Projection}
\label{app:projection_theory}

The final prompt update uses \(S_r^{(T-1)}\), the final response
update produces \(S_r^{(T)}\), and Closure produces
\(S_r^{\mathrm{cl}}\) while retaining \(S_p^{(T)}\).
For one example, suppress \(k\), set \(V_0=S_r^{(T-1)}\) and
\(V_{\mathrm{cl}}=S_r^{\mathrm{cl}}\), and define
\[
s_i^0=\Delta_p(i\mid V_0),
\qquad
d_i=\Delta_p(i\mid V_{\mathrm{cl}})-s_i^0.
\]
For \(1\le b_p<|\mathcal P|\), let
\[
\begin{aligned}
\mu_p^0&=
\min_{\substack{i\in S_p^{(T)}\\
i'\in\mathcal P\setminus S_p^{(T)}}}
(s_i^0-s_{i'}^0),\\
\varepsilon_{\mathrm{cl}}&=\max_{i\in\mathcal P}|d_i|.
\end{aligned}
\]

\begin{proposition}[Prompt stability after response projection]
\label{prop:projection_prompt_stability}
If \(\mu_p^0>2\varepsilon_{\mathrm{cl}}\), then
\(S_p^{(T)}\) remains a Top-\(b_p\) subset under
\(V_{\mathrm{cl}}\).
\end{proposition}

\begin{proof}
Each selected--unselected score difference under the projected state
is at least \(\mu_p^0-2\varepsilon_{\mathrm{cl}}>0\).
\end{proof}

The score change is computed from cached interactions:
\begin{equation}
\label{eq:projection_prompt_score_change}
\begin{aligned}
d_i={}&
\sum_{j\in V_{\mathrm{cl}}\setminus V_0}\widetilde\beta_{i,j}\\
&-\sum_{j\in V_0\setminus V_{\mathrm{cl}}}\widetilde\beta_{i,j}.
\end{aligned}
\end{equation}
The condition permits different response-set cardinalities.

To distinguish terminal reconditioning from projection, define
\[
\begin{aligned}
S_p^{\mathrm{pre}}
&=\operatorname{Top}_{b_p}\Delta_p(\cdot\mid S_r^{(T)}),\\
S_p^{\mathrm{post}}
&=\operatorname{Top}_{b_p}\Delta_p(\cdot\mid S_r^{\mathrm{cl}}).
\end{aligned}
\]
Comparing \(S_p^{(T)}\) with \(S_p^{\mathrm{pre}}\) measures the
effect of the terminal response update; comparing
\(S_p^{\mathrm{pre}}\) with \(S_p^{\mathrm{post}}\) isolates the
additional effect of Closure. The Post-Closure refresh experiment
evaluates the combined update from \(S_p^{(T)}\) to
\(S_p^{\mathrm{post}}\).
Budget feasibility follows the cap-enforcing projection rules described
in Appendix~\ref{app:structural_projection}.

\section{Experimental Details}
\label{app:implementation}
This section describes the experimental protocol shared by the main and
supplementary analyses. Results in the main tables and the
forgetting-mitigation comparison are reported across three random seeds.
Supplementary analyses of mechanisms and design choices use single runs
unless otherwise specified in the corresponding table or figure.

\subsection{Data and Evaluation}
\label{app:data_split}
\label{app:metrics_details}
We use a fixed partition of MetaMathQA-40K with controls for data leakage
across all methods and backbones. After removing problem groups that overlap
with the MATH evaluation set, we partition the resulting pool of
39{,}973 examples at the problem-group level into three pairwise-disjoint
subsets:
\begin{equation}
\begin{aligned}
D_{\mathrm{all}} &= D_{\mathrm{w}}\,\dot\cup\,D_{\mathrm{val}}\,\dot\cup\,D_{\mathrm{sft}}, \\
D_a\cap D_b &= \varnothing\quad \text{for}\ a\neq b,
\end{aligned}
\end{equation}
with \(|D_{\mathrm{w}}|=2{,}004\) (approximately 5\%),
\(|D_{\mathrm{val}}|=1{,}963\) (approximately 5\%), and
\(|D_{\mathrm{sft}}|=36{,}006\) (approximately 90\%). We generate this partition once using
split seed 42 and reuse it across all methods, backbones, and reported
training seeds. The label ``5\% warmup'' denotes the approximately
5\% warmup subset specified above.

\begin{table}[t]
\centering
\small
\setlength{\tabcolsep}{3.5pt}
\begin{tabular}{lccc}
\toprule
\textbf{Split} & \textbf{Size} & \textbf{Ratio} & \textbf{Role} \\
\midrule
$D_{\mathrm{w}}$   & 2{,}004  & 5.01\%  & Warmup checkpoint \\
$D_{\mathrm{val}}$ & 1{,}963  & 4.91\%  & Validation direction \\
$D_{\mathrm{sft}}$ & 36{,}006 & 90.08\% & Scoring, selection, and final SFT \\
\bottomrule
\end{tabular}
\vspace{-0.1in}
\caption{MetaMathQA-40K data partition used throughout the main experiments. Ratios are computed over the 39{,}973-example pool and rounded to two decimal places.}
\label{tab:data_split}
\vspace{-0.2in}
\end{table}

The warmup split $D_{\mathrm{w}}$ is used exclusively to obtain the scoring
checkpoint $\bar{\theta}$ through one epoch of standard SFT. The held-out
split $D_{\mathrm{val}}$ is used exclusively to compute the target validation
direction
$g_{\mathrm{target}}=\nabla_\theta L_{\mathrm{val}}(\bar{\theta};D_{\mathrm{val}})$;
its examples are never used for optimization during warmup or final SFT.
Token scoring, selection, and compact instruction tuning are performed on
$D_{\mathrm{sft}}$, which corresponds to $D$ in Section~\ref{sec:method} and
Algorithm~\ref{alg:BRIDGE}. Final SFT always restarts from the common
pretrained initialization $\theta_{\mathrm{old}}$ rather than continuing
from $\bar{\theta}$. Thus, neither warmup nor validation examples contribute
directly to the final SFT objective.

\paragraph{Evaluation.}
To evaluate general capabilities, we report raw accuracy on
ARC-C~\citep{clark2018think} and HellaSwag~\citep{zellers2019hellaswag},
5-shot accuracy on MMLU~\citep{hendrycks2021measuring}, and pass@1 on
HumanEval~\citep{chen2021evaluating}, using the native code-fence extraction
procedure for each model. To evaluate mathematical reasoning, we report
numeric accuracy on GSM8K~\citep{cobbe2021training} by comparing the extracted
final number with the reference answer after normalization. We report
equivalence-based accuracy on MATH~\citep{hendrycks2021measuringa} by
comparing parsed expressions after LaTeX normalization.

We define the aggregate metrics used throughout the paper as follows.
$\operatorname{GeneralAvg}$ is the unweighted mean of the scores on ARC-C,
HellaSwag, MMLU, and HumanEval. $\operatorname{OverallAvg}$ is the unweighted
mean of $\operatorname{GeneralAvg}$ and the score on the target-task
benchmark. $\operatorname{Forget}*{\mathrm{general}}$ is the decrease in
$\operatorname{GeneralAvg}$ relative to the pretrained checkpoint, and
$\operatorname{Gain}*{\mathrm{target}}$ is the improvement on the target-task
benchmark achieved by the fine-tuned model relative to the pretrained
checkpoint. We also report the retained prompt- and response-token ratios
$\rho_p=|S_p|/|P_k|$ and $\rho_r=|S_r|/|R_k|$, together with the number of
directly supervised response positions $|S_r|$.

\subsection{Training and Baseline Configurations}
\label{app:training_config}
\label{app:baseline_details}
BRIDGE comprises three main stages: (1) warmup training, (2) token scoring
and selection, and (3) compact instruction tuning.
Table~\ref{tab:training_config} summarizes the BRIDGE training configuration.
Table~\ref{tab:hyperparams} reports the selected learning rates and epoch
counts for all selection methods. For each method--backbone pair, the
selected configuration is the same across MetaMathQA-40K, Magicoder,
and Alpaca.

\begin{table}[t]
\centering
\small
\resizebox{\columnwidth}{!}{%
\begin{tabular}{lc}
\toprule
\textbf{Configuration} & \textbf{Value} \\
\midrule
Model & \makecell[c]{Llama-3.2-3B / Gemma-3-4B /\\ Qwen3.5-9B} \\
Optimizer & AdamW \\
Learning Rate & $2\times10^{-5}$, cosine \\
Batch Size & 64 (effective) \\
Gradient Accumulation & 8 \\
Training Epochs & 3 \\
Maximum Sequence Length & \makecell[c]{1{,}024 $+$ 1{,}024\\ (prompt $+$ response)} \\
Precision & BF16 \\
\bottomrule
\end{tabular}%
}
\vspace{-0.1in}
\caption{
Training configurations used for BRIDGE.
}
\label{tab:training_config}
\vspace{-0.2in}
\end{table}

\begin{table*}[t]
\centering
\small
\setlength{\tabcolsep}{6pt}
\begin{tabular*}{\textwidth}{@{\extracolsep{\fill}}lccc@{}}
\toprule
\textbf{Method} & \textbf{Llama-3.2-3B} & \textbf{Gemma-3-4B} & \textbf{Qwen3.5-9B} \\
\midrule
Full SFT       & $2\times10^{-5}\,/\,3$ & $2\times10^{-5}\,/\,3$ & $2\times10^{-5}\,/\,3$ \\
Random         & $2\times10^{-5}\,/\,3$ & $2\times10^{-5}\,/\,3$ & $1\times10^{-5}\,/\,2$ \\
RHO-1          & $2\times10^{-5}\,/\,3$ & $2\times10^{-5}\,/\,3$ & $2\times10^{-5}\,/\,3$ \\
LLMLingua-2    & $2\times10^{-5}\,/\,3$ & $2\times10^{-5}\,/\,3$ & $1\times10^{-5}\,/\,2$ \\
ssToken        & $2\times10^{-5}\,/\,3$ & $2\times10^{-5}\,/\,3$ & $1\times10^{-4}\,/\,1$ \\
TokenSeek      & $2\times10^{-5}\,/\,3$ & $2\times10^{-5}\,/\,3$ & $2\times10^{-5}\,/\,3$ \\
Token Cleaning & $2\times10^{-5}\,/\,3$ & $2\times10^{-5}\,/\,3$ & $2\times10^{-5}\,/\,3$ \\
LLM+TC       & $2\times10^{-5}\,/\,3$ & $2\times10^{-5}\,/\,3$ & $1\times10^{-5}\,/\,2$ \\
BRIDGE         & $2\times10^{-5}\,/\,3$ & $2\times10^{-5}\,/\,3$ & $2\times10^{-5}\,/\,3$ \\
\bottomrule
\end{tabular*}
\vspace{-0.1in}
\caption{Selected optimization hyperparameters. Each cell reports learning rate / training epochs. The same configuration is used on MetaMathQA-40K, Magicoder, and Alpaca for each method--backbone pair.}
\label{tab:hyperparams}
\end{table*}

During warmup training, BRIDGE obtains a reference checkpoint
$\bar{\theta}$ for subsequent token-level scoring through standard
instruction tuning without token selection. During compact instruction
tuning, BRIDGE uses the selected prompt tokens as the reduced context and
computes the training loss only at selected response positions. The
remaining response tokens are retained as autoregressive history to
maintain generation consistency.

\paragraph{Baseline implementations and budget matching.}
For each backbone and dataset, the selection baselines use the same data
splits, tokenizer, optimizer, scheduler, and evaluation protocol as BRIDGE.
Their learning rates and epoch counts are method-specific
(Table~\ref{tab:hyperparams}), while token-selection budgets follow
the matching protocol described below.

\emph{Pretrained} evaluates the original checkpoint without fine-tuning.
\emph{Standard SFT} fine-tunes on all prompt and response tokens without
selection. \emph{Random} retains a uniformly sampled token subset with the
same retention ratios as BRIDGE. \emph{LLM+TC(Ind.)} replaces the coupled
selection objective with two independent single-end selectors, one for
prompts and one for responses, each operating under the corresponding
matched budget (Section~\ref{subsec:coupled_selection}). \emph{Span Random}
samples random contiguous response spans under the matched budget
(Section~\ref{subsec:structure_projection}).

For external selection methods, we follow the released method descriptions
and adapt their original configurations to the matched-budget protocol
where necessary. RHO-1~\citep{lin2024not} selects tokens based on excess loss
relative to a reference model and computes the training loss only at
selected positions. LLMLingua-2~\citep{pan2024llmlingua} removes prompt tokens
using a token-classification compressor, whereas
ssToken~\citep{qin2026sstoken}, TokenSeek~\citep{zeng2026tokenseek}, and
Token Cleaning~\citep{pang2025token} select or filter instruction-tuning
tokens using their respective scoring mechanisms. Dual-end methods use
the same prompt and response retention ratios as BRIDGE,
$(\rho_p,\rho_r)=(0.75,0.75)$. Single-side baselines retain all tokens on the
side where selection is not applied and adjust the budget on the selectable
side to match the total retained-token count of BRIDGE as closely as
feasible:
\begin{equation}
B_k^{\mathrm{match}}
=
0.75|P_k|+0.75|R_k|.
\end{equation}
Residual mismatches occur only when the target total cannot be achieved
within the selection scope of a single-side method or under integer token
constraints.

\paragraph{Realized two-sided budgets.}
We compute the following totals from the saved selection masks and the
original scored sequence lengths for the 36,006-example SFT split.
Retention rates exclude fixed BOS and EOS positions. Executed token counts
include the full response and exclude batch padding.
\begin{table}[t]
\centering
\small
\setlength{\tabcolsep}{1.4pt}
\begin{tabular}{lccc}
\toprule
\textbf{Model} &
\textbf{Prompt} &
\textbf{Response loss} &
\textbf{Executed / full} \\
\midrule
Llama  & 75.00\% & 75.00\% & 7.893 / 8.552 M \\
Gemma  & 75.01\% & 75.00\% & 8.602 / 9.285 M \\
Qwen  & 75.00\% & 75.00\% & 8.680 / 9.378 M \\
\bottomrule
\end{tabular}
\vspace{-0.1in}
\caption{Retention rates for eligible positions and unpadded executed token
counts per pass through the SFT split in the saved runs. Closure uses
$97\%$ of its eligible response budget in these runs.}
\label{tab:realized_two_sided_budgets}
\end{table}

\subsection{Scoring and Structural Projection}
\label{app:scoring_config}

The scoring stage constructs a gradient signal directed by the validation
split and estimates prompt--response interactions under fixed prompt and
response budgets. Table~\ref{tab:scoring_config} summarizes the default
configuration used throughout the main experiments.

\begin{table}[t]
\centering
\small
\resizebox{\columnwidth}{!}{%
\begin{tabular}{lc}
\toprule
\textbf{Configuration} & \textbf{Value} \\
\midrule
Gradient Direction & Validation-split direction \\
Gradient Scope & Final four Transformer layers \\
Retention Anchor & \makecell[c]{L2 parameter-space anchor\\ $\lambda=0.2$} \\
Attention Layer & \makecell[c]{Llama: final Transformer layer\\ Gemma: last global-attention layer (29)\\ Qwen: last full-attention layer (31)} \\
Interaction Surrogate & $A_{k,j,i},a_{k,j}$ \\
Warmup Ratio & $\approx 5\%$ \\
Prompt Budget & 75\% \\
Response Budget & 75\% \\
Alternating Iterations & $T_{\mathrm{alt}}=4$ \\
Structural Projection & Deterministic Closure \\
\bottomrule
\end{tabular}%
}
\vspace{-0.1in}
\caption{
Default BRIDGE scoring and selection settings. The L2 anchor affects
scoring only and is not used in final SFT.
}
\label{tab:scoring_config}
\vspace{-0.2in}
\end{table}

For each example, BRIDGE first computes response-token utilities directed
by the validation split, using gradients restricted to the final four
Transformer layers. To incorporate a lightweight retention prior without
an additional retention dataset, we combine the target validation direction
with an L2 parameter-space anchor defined by the displacement from the
pretrained checkpoint $\theta_{\mathrm{old}}$ to the warmup checkpoint $\bar{\theta}$.
The anchor direction is $\bar{\theta}-\theta_{\mathrm{old}}$ and is combined with the
target direction using $\lambda=0.2$. This anchor affects only token
valuation. Final compact SFT uses the original training objective and
restarts from the pretrained initialization.

Response-to-prompt attention then propagates response-token utility to
prompt positions through the interaction surrogate

$$
\widetilde{\beta}_{k,i,j}=A_{k,j,i}a_{k,j}.
$$

Attention routing uses the final Transformer layer for Llama, the last
global-attention layer (layer 29) for Gemma, and the last full-attention
layer (layer 31) for Qwen. These attention layers determine the routing
weights; gradient scoring uses the final four Transformer layers.

Routing weights are obtained from the query row at the input position of
$r_{k,j}$, rather than the preceding position whose logits predict
$r_{k,j}$. Each head applies a single softmax over all unmasked causal
keys, including the current token itself, before weights are averaged
across heads. Prompt and response-prefix weights are extracted without
renormalization; the history term sums only positions $j'<j$, excluding
self-attention.

BRIDGE alternates between prompt and response selection for
$T_{\mathrm{alt}}=4$ rounds under fixed retention ratios
$\rho_p=\rho_r=0.75$. After the final token-level update, a deterministic
structure-aware projection is applied exclusively to the response
selection. This projection reallocates the same response-supervision
budget across coherent local units without additional model inference or
gradient computation. The resulting static response mask is used during
final SFT. The selected prompt is physically compacted, while the full
response is retained as autoregressive history. Detailed projection rules
and diagnostics are provided in Appendix~\ref{app:closure_diagnostics}.

\section{Selection-Conditioned Utility}
\label{app:selection_conditioned_utility}
This section presents additional evidence for selection-conditioned token
utility, extending the analysis in Section~\ref{sec:coupling_analysis}. We
first examine how selection on one side changes token rankings on the
other, then define controlled selection schedules within the shared BRIDGE
scoring formulation. Appendix~\ref{app:alternating_rounds} reports
sensitivity to the number of alternating rounds, and
Appendix~\ref{app:baseline_combinations} examines combinations of existing
selectors.

\subsection{Cross-Side Utility Changes}
\label{app:cross_side_utility}
The analysis in Fig.~\ref{fig:coupling_analysis}(b) directly measures how
token rankings change as the retained subset on the opposite side varies.
For prompt tokens, we recompute conditional scores over progressively
smaller sets of response tokens retained for supervision and compare each
resulting ranking with the ranking obtained under full response retention.
For response tokens, we conduct the corresponding analysis by varying the
retained prompt context. We measure ranking agreement using Spearman rank
correlation. As retention on the opposite side decreases, agreement with
the full-retention ranking declines, reaching $0.34$ at the lowest retention
level evaluated. These results support modeling token utility as
selection-conditioned rather than assigning each token a fixed score.

\subsection{Controlled Selection Schedules}
\label{app:coupling_schedule}
\label{app:coupling_definitions}
To isolate the effect of mutual conditioning, we keep the precomputed
interaction scores $\widetilde{\beta}$, history utilities
$\widetilde{\gamma}$, token budgets, closure projection, and final training
configuration fixed across all schedules. The schedules differ only in
how they construct the prompt and response subsets.

\paragraph{Independent (reported as $T=0$).}
Each side is selected once, conditioned on the full opposite side:
\begin{equation}
\label{eq:independent_selection}
\begin{aligned}
S_{p,k}^{\mathrm{ind}}
&=\operatorname{Top}_{b_{p,k}}
\bigl\{\Delta_{p,k}(i\mid R_k)\bigr\}_{i\in P_k},\\
S_{r,k}^{\mathrm{ind}}
&=\operatorname{Top}_{b_{r,k}}
\bigl\{\Delta_{r,k}(j\mid P_k)\bigr\}_{j\in R_k}.
\end{aligned}
\end{equation}
Both rankings are computed from the original full prompt--response pair,
with no rescoring after selection on the opposite side. In the plots,
$T=0$ denotes this non-iterative reference, defined by
Eq.~\eqref{eq:independent_selection}, \emph{not} a zero-round execution of
Algorithm~\ref{alg:BRIDGE}.

\paragraph{One-Pass ($T=1$).}
One-Pass executes a single alternating round of
Algorithm~\ref{alg:BRIDGE}. It first initializes the response subset using
history utility:
\begin{equation}
S_{r,k}^{(0)}
=
\operatorname{Top}_{b_{r,k}}
\bigl\{\widetilde{\gamma}_{k,j}\bigr\}_{j\in R_k},
\end{equation}
then updates the prompt subset followed by the response subset:
\begin{equation}
\label{eq:one_pass_selection}
\begin{aligned}
S_{p,k}^{(1)}
&=\operatorname{Top}_{b_{p,k}}
\bigl\{\Delta_{p,k}(i\mid S_{r,k}^{(0)})\bigr\}_{i\in P_k},\\
S_{r,k}^{(1)}
&=\operatorname{Top}_{b_{r,k}}
\bigl\{\Delta_{r,k}(j\mid S_{p,k}^{(1)})\bigr\}_{j\in R_k}.
\end{aligned}
\end{equation}
The prompt ranking is conditioned on the response subset selected under
the response budget. The response ranking is then conditioned on the
newly selected prompt subset. These updates complete the selection round,
after which closure projection is applied using response scores
conditioned on $S_{p,k}^{(1)}$.

\paragraph{Alternating-$T$.}
For $T>1$, BRIDGE repeats the two updates in
Eq.~\eqref{eq:alternating_updates} for $T$ rounds. Each update selects a new
subset conditioned on the current subset on the opposite side. The default
configuration uses $T=4$.

\section{Additional Results and Capability Retention}
\label{app:additional_tasks}
This section extends the effectiveness and capability-retention results
presented in Section~\ref{subsec:main_results}. We first compare independent
and sequential combinations of existing selectors, then evaluate code
generation and instruction following, and finally compare BRIDGE with
methods designed to mitigate forgetting.

\subsection{Independent and Sequential Combinations of Existing Selectors}
\label{app:baseline_combinations}
\label{app:independent_combinations}

The analysis in Appendix~\ref{app:coupling_schedule} compares update
schedules within the BRIDGE token-utility formulation. Here, we examine
whether combinations of existing single-side selectors provide comparable
benefits. These experiments distinguish the effect of combining prompt
and response selectors under matched budgets from that of conditioning
token utility on the subset retained on the opposite side.

We conduct this comparison on Llama-3.2-3B using the same MetaMathQA split,
training initialization, optimizer, scheduler, and evaluation protocol as
the main experiments. Unless otherwise specified, all combined methods
retain $75\%$ of eligible prompt tokens and supervise $75\%$ of eligible
response tokens. The final training procedure is also held fixed: the
selected prompt is physically compacted, the complete response remains
available as causal history, and the loss is applied only at selected
response positions. The methods differ only in how they construct the
prompt and response subsets.

\paragraph{Independent composition.}
We first combine prompt-side and response-side selectors without
cross-side rescoring. LLMLingua-2 selects prompt tokens, while RHO-1,
ssToken, and Token Cleaning serve as alternative response-side selectors.

For an example $(P_k,R_k)$, let the prompt-side method $f_p$ and
response-side method $f_r$ produce
\begin{equation}
\begin{aligned}
S_{p,k}^{f_p}
&=
f_p(P_k,R_k;\rho_p), \\
S_{r,k}^{f_r}
&=
f_r(P_k,R_k;\rho_r).
\end{aligned}
\label{eq:independent_baseline_composition}
\end{equation}
Both selectors operate on the original, uncompressed prompt--response
pair. Their outputs are combined when constructing the final training
example. Response selection therefore remains conditioned on the full
prompt, while prompt selection remains independent of the response
positions retained for supervision.

We evaluate three combinations:
\emph{LLM+RHO-1 (Ind.)},
\emph{LLM+ssToken (Ind.)}, and
\emph{LLM+TC (Ind.)}.
Here, ``LLM'' denotes LLMLingua-2 prompt compression, and ``TC'' denotes
Token Cleaning. The previously reported LLM+TC baseline corresponds to
this independent-composition setting. These controls assess the
performance of existing specialized selectors when prompt and response
selection are performed independently.

\paragraph{Sequential one-way conditioning.}
\label{app:sequential_baseline}
We next introduce \emph{LLM$\rightarrow$TC (Seq.)}, which conditions
response selection on the compressed prompt used for final training.

LLMLingua-2 first selects the prompt subset,
\begin{equation}
S_{p,k}^{\mathrm{LLM}}
=
\operatorname{LLMLingua}
(P_k,R_k;\rho_p),
\label{eq:sequential_llm_prompt}
\end{equation}
which is then physically compacted:
\begin{equation}
\widetilde{P}_k^{\mathrm{LLM}}
=
\operatorname{Compact}
\left(
P_k,
S_{p,k}^{\mathrm{LLM}}
\right).
\label{eq:sequential_llm_compact}
\end{equation}
Token Cleaning is subsequently rerun on the compressed example:
\begin{equation}
S_{r,k}^{\mathrm{TC}\mid\mathrm{LLM}}
=
\operatorname{TokenCleaning}
\left(
\widetilde{P}_k^{\mathrm{LLM}},
R_k;
\rho_r
\right).
\label{eq:sequential_tc_response}
\end{equation}
The resulting training example uses
$\widetilde{P}_k^{\mathrm{LLM}} \mathbin{\Vert} R_k$ as model input, with
the training loss applied only at positions in
$S_{r,k}^{\mathrm{TC}\mid\mathrm{LLM}}$.
Response selection is thus evaluated under the same prompt context used
during final SFT.

This procedure introduces a single dependency
$S_p \longrightarrow S_r$:
the prompt subset remains fixed after response selection. BRIDGE instead
uses the repeated mutual dependency
$S_r^{(t)} \longrightarrow S_p^{(t+1)} \longrightarrow S_r^{(t+1)}$,
updating each subset according to the subset retained on the opposite
side. The sequential baseline evaluates the benefit of conditioning
response selection on prompt compression, complementing the analysis of
repeated mutual conditioning in
Appendix~\ref{app:coupling_schedule}.

\paragraph{Comparison.}
\label{app:baseline_combination_comparison}
Table~\ref{tab:baseline_combinations} compares three forms of cross-side
dependency. Independent composition selects both sides using the original
example. Sequential conditioning selects response tokens using the
compressed prompt. BRIDGE repeatedly updates prompt and response
surrogate utilities by aggregating precomputed interactions over the
currently retained subset on the opposite side.

\begin{table*}[t]
\centering
\small
\setlength{\tabcolsep}{4.5pt}
\resizebox{\textwidth}{!}{%
\begin{tabular}{llllccc}
\toprule
\textbf{Method}
&
\textbf{Prompt Selector}
&
\textbf{Response Selector}
&
\textbf{Cross-Side Dependency}
&
\textbf{MATH}
&
\textbf{General Avg.}
&
\textbf{Overall Avg.} \\
\midrule
LLM+RHO-1 (Ind.) & LLMLingua-2 & RHO-1 & Independent & 12.1 & 45.81 & 28.96 \\
LLM+ssToken (Ind.) & LLMLingua-2 & ssToken & Independent & 12.34 & 46.02 & 29.18 \\
LLM+TC (Ind.) & LLMLingua-2 & Token Cleaning & Independent & 13.32 & 45.57 & 29.44 \\
LLM$\rightarrow$TC (Seq.) & LLMLingua-2 & Token Cleaning & Prompt $\rightarrow$ Response & 13.52 & 46.06 & 29.79 \\
\rowcolor{oursrow}\cellcolor{white}
\textbf{BRIDGE}\,\venuetag{Ours} & \textbf{BRIDGE} & \textbf{BRIDGE} & \textbf{Mutual, alternating} & 15.96 & 46.92 & 31.44 \\
\bottomrule
\end{tabular}%
}
\vspace{-0.1in}
\caption{
Independent, sequential, and alternating prompt--response selection on
Llama-3.2-3B with matched retention ratios and final SFT settings.
}
\label{tab:baseline_combinations}
\end{table*}

With the same pair of external selectors, rerunning Token Cleaning after
LLMLingua-2 prompt compression improves MATH accuracy from $13.32$ to
$13.52$ and Overall Avg.\ from $29.44$ to $29.79$. This improvement
indicates that response selection benefits from the prompt context used
for final training. BRIDGE achieves $15.96$ on MATH and $31.44$ in Overall
Avg. Together, these experiments compare the performance of independent,
sequential, and mutually conditioned selection. The schedule analysis in
Appendix~\ref{app:coupling_schedule} further isolates the effect of
repeated conditioning within the shared BRIDGE scoring formulation.

\subsection{Code Generation}
\label{app:code_generation}
The code-generation and instruction-following experiments share the
following protocol. Unless otherwise stated, both follow the
mathematical-reasoning protocol: each fine-tuning corpus is divided into
pairwise-disjoint subsets with target proportions of 5\% for warmup,
5\% for validation, and 90\% for final SFT. Final training restarts from the pretrained
initialization. The method- and backbone-specific learning rates and epoch
counts in Table~\ref{tab:hyperparams} also apply to both datasets.
The optimizer, effective batch size, scoring configuration, token budgets,
and alternating-selection settings remain unchanged.

For code generation, models are fine-tuned on
Magicoder~\citep{wei2024magicoder} and evaluated on
HumanEval~\citep{chen2021evaluating} using pass@1. Code statements serve
as the structural units for closure, following recommendations for
code supervision at appropriate block
granularities~\citep{deng2026codeblock}.

\begin{table*}[t]
\centering
\small
\setlength{\tabcolsep}{4.5pt}
\renewcommand{\arraystretch}{0.85}
\setlength{\aboverulesep}{1.2pt}
\setlength{\belowrulesep}{1.2pt}

\resizebox{\textwidth}{!}{%
\begin{tabular}{llccccccc}
\toprule
\multicolumn{1}{c}{\multirow{2}{*}{\textbf{Model}}} & \multirow{2}{*}{\textbf{Method}} & \multicolumn{5}{c}{\textbf{General Capability (\%)}} & \multicolumn{1}{c}{\textbf{Target (\%)}} & \multicolumn{1}{c}{\textbf{Overall}} \\
\cmidrule(lr){3-7} \cmidrule(lr){8-8} \cmidrule(lr){9-9}
& & \textbf{ARC-C} & \textbf{HellaSwag} & \textbf{MMLU} & \textbf{GSM8K} & \textbf{Avg.} & \textbf{HumanEval} & \textbf{Avg.} \\
\midrule

\multirow{10}{*}{\rotatebox{90}{\textbf{Llama-3.2-3B}}}
& \color{baselinecolor} Pretrained
& \color{baselinecolor} 43.17
& \color{baselinecolor} 54.20
& \color{baselinecolor} 56.64
& \color{baselinecolor} 28.28
& \color{baselinecolor} 45.57
& \color{baselinecolor} 28.05
& \color{baselinecolor} 36.81
\\

& \color{baselinecolor} Full SFT
& \color{baselinecolor} $39.96_{\pm 0.26}$
& \color{baselinecolor} $52.45_{\pm 0.27}$
& \color{baselinecolor} $48.23_{\pm 0.19}$
& \color{baselinecolor} $29.11_{\pm 0.57}$
& \color{baselinecolor} $42.44_{\pm 0.24}$
& \color{baselinecolor} $44.72_{\pm 0.93}$
& \color{baselinecolor} $43.58_{\pm 0.39}$
\\

& Random
& $40.73_{\pm 0.69}$
& $53.36_{\pm 0.46}$
& $53.65_{\pm 0.41}$
& $25.45_{\pm 0.44}$
& $43.30_{\pm 0.36}$
& $35.37_{\pm 1.06}$
& $39.34_{\pm 0.39}$
\\

& RHO-1\,\venuetag{NeurIPS'24}
& $41.33_{\pm 0.26}$
& \second{$54.07_{\pm 0.11}$}
& \best{$55.93_{\pm 0.36}$}
& $24.92_{\pm 0.29}$
& $44.06_{\pm 0.19}$
& $35.37_{\pm 0.61}$
& $39.72_{\pm 0.21}$
\\

& LLMLingua-2\,\venuetag{ACL'24}
& $41.98_{\pm 0.52}$
& $52.80_{\pm 0.29}$
& $53.32_{\pm 0.32}$
& $25.45_{\pm 0.44}$
& $43.39_{\pm 0.28}$
& \second{$41.87_{\pm 1.27}$}
& $42.63_{\pm 0.51}$
\\

& ssToken\,\venuetag{ICLR'26}
& $43.83_{\pm 0.52}$
& $53.56_{\pm 0.33}$
& $52.90_{\pm 0.29}$
& $25.65_{\pm 0.29}$
& $43.99_{\pm 0.26}$
& $41.67_{\pm 1.76}$
& \second{$42.83_{\pm 0.77}$}
\\

& TokenSeek\,\venuetag{ICLR'26}
& $42.21_{\pm 0.66}$
& $53.30_{\pm 0.31}$
& $54.69_{\pm 0.31}$
& \second{$26.36_{\pm 0.49}$}
& \second{$44.14_{\pm 0.32}$}
& $40.04_{\pm 1.27}$
& $42.09_{\pm 0.48}$
\\

& TokenCleaning\,\venuetag{ICML'25}
& \best{$44.62_{\pm 0.37}$}
& $52.80_{\pm 0.29}$
& $53.62_{\pm 0.35}$
& $24.24_{\pm 0.53}$
& $43.82_{\pm 0.28}$
& $41.06_{\pm 0.35}$
& $42.44_{\pm 0.25}$
\\

& LLM+TC (Ind.)
& $43.29_{\pm 0.20}$
& $53.51_{\pm 0.30}$
& $53.21_{\pm 0.33}$
& $24.21_{\pm 0.46}$
& $43.55_{\pm 0.23}$
& $40.45_{\pm 0.35}$
& $42.00_{\pm 0.21}$
\\

\rowcolor{oursrow}
\cellcolor{white}
& \textbf{BRIDGE}\,\venuetag{Ours}
& \second{$44.51_{\pm 0.27}$}
& \best{$54.48_{\pm 0.16}$}
& \second{$55.19_{\pm 0.35}$}
& \best{$29.21_{\pm 0.46}$}
& \best{$45.85_{\pm 0.23}$}
& \best{$43.09_{\pm 0.93}$}
& \best{$44.47_{\pm 0.37}$}
\\

\midrule

\multirow{10}{*}{\rotatebox{90}{\textbf{Gemma-3-4B-PT}}}
& \color{baselinecolor} Pretrained
& \color{baselinecolor} 51.37
& \color{baselinecolor} 54.86
& \color{baselinecolor} 59.72
& \color{baselinecolor} 44.28
& \color{baselinecolor} 52.56
& \color{baselinecolor} 35.37
& \color{baselinecolor} 43.96
\\

& \color{baselinecolor} Full SFT
& \color{baselinecolor} $55.66_{\pm 0.18}$
& \color{baselinecolor} $57.30_{\pm 0.37}$
& \color{baselinecolor} $59.60_{\pm 0.33}$
& \color{baselinecolor} $42.99_{\pm 1.14}$
& \color{baselinecolor} $53.89_{\pm 0.39}$
& \color{baselinecolor} $55.08_{\pm 0.93}$
& \color{baselinecolor} $54.49_{\pm 0.39}$
\\

& Random
& $54.35_{\pm 0.26}$
& $56.36_{\pm 0.35}$
& $58.89_{\pm 0.10}$
& $40.28_{\pm 0.61}$
& $52.47_{\pm 0.25}$
& $47.97_{\pm 1.88}$
& $50.22_{\pm 0.82}$
\\

& RHO-1\,\venuetag{NeurIPS'24}
& $53.87_{\pm 0.34}$
& $54.77_{\pm 0.24}$
& $57.66_{\pm 0.20}$
& $36.92_{\pm 0.53}$
& $50.81_{\pm 0.24}$
& $52.24_{\pm 0.39}$
& $51.53_{\pm 0.22}$
\\

& LLMLingua-2\,\venuetag{ACL'24}
& \best{$55.92_{\pm 0.30}$}
& $56.33_{\pm 0.22}$
& $58.79_{\pm 0.25}$
& $42.03_{\pm 0.57}$
& \second{$53.27_{\pm 0.25}$}
& $50.41_{\pm 2.46}$
& $51.84_{\pm 1.13}$
\\

& ssToken\,\venuetag{ICLR'26}
& $53.56_{\pm 0.64}$
& \best{$57.27_{\pm 0.14}$}
& $59.27_{\pm 0.28}$
& \second{$42.38_{\pm 0.45}$}
& $53.12_{\pm 0.28}$
& $54.07_{\pm 1.14}$
& \second{$53.60_{\pm 0.45}$}
\\

& TokenSeek\,\venuetag{ICLR'26}
& $54.69_{\pm 0.37}$
& $54.98_{\pm 0.31}$
& \second{$59.39_{\pm 0.31}$}
& $41.19_{\pm 0.63}$
& $52.56_{\pm 0.29}$
& $53.25_{\pm 1.19}$
& $52.91_{\pm 0.47}$
\\

& TokenCleaning\,\venuetag{ICML'25}
& $53.38_{\pm 0.43}$
& $55.05_{\pm 0.25}$
& $57.56_{\pm 0.25}$
& $35.89_{\pm 0.46}$
& $50.47_{\pm 0.25}$
& \second{$54.27_{\pm 1.32}$}
& $52.37_{\pm 0.56}$
\\

& LLM+TC (Ind.)
& $53.87_{\pm 0.64}$
& $54.54_{\pm 0.24}$
& $58.24_{\pm 0.27}$
& $39.83_{\pm 0.87}$
& $51.62_{\pm 0.38}$
& $51.42_{\pm 1.37}$
& $51.52_{\pm 0.55}$
\\

\rowcolor{oursrow}
\cellcolor{white}
& \textbf{BRIDGE}\,\venuetag{Ours}
& \second{$54.84_{\pm 0.34}$}
& \second{$56.68_{\pm 0.15}$}
& \best{$59.57_{\pm 0.27}$}
& \best{$44.66_{\pm 0.53}$}
& \best{$53.94_{\pm 0.24}$}
& \best{$56.71_{\pm 0.61}$}
& \best{$55.33_{\pm 0.25}$}
\\

\midrule

\multirow{10}{*}{\rotatebox{90}{\textbf{Qwen3.5-9B}}}
& \color{baselinecolor} Pretrained
& \color{baselinecolor} 54.27
& \color{baselinecolor} 54.26
& \color{baselinecolor} 72.01
& \color{baselinecolor} 89.08
& \color{baselinecolor} 67.41
& \color{baselinecolor} 61.59
& \color{baselinecolor} 64.50
\\

& \color{baselinecolor} Full SFT
& \color{baselinecolor} $50.94_{\pm 0.26}$
& \color{baselinecolor} $52.35_{\pm 0.31}$
& \color{baselinecolor} $72.54_{\pm 0.22}$
& \color{baselinecolor} $87.29_{\pm 0.64}$
& \color{baselinecolor} $65.78_{\pm 0.26}$
& \color{baselinecolor} $72.76_{\pm 1.53}$
& \color{baselinecolor} $69.27_{\pm 0.71}$
\\

& Random
& $52.13_{\pm 0.34}$
& $53.27_{\pm 0.23}$
& $71.78_{\pm 0.22}$
& $86.71_{\pm 0.68}$
& $65.97_{\pm 0.27}$
& $66.87_{\pm 1.27}$
& $66.42_{\pm 0.52}$
\\

& RHO-1\,\venuetag{NeurIPS'24}
& $52.10_{\pm 0.34}$
& $52.95_{\pm 0.34}$
& $71.48_{\pm 0.17}$
& \second{$87.67_{\pm 0.38}$}
& $66.05_{\pm 0.22}$
& \second{$69.72_{\pm 1.86}$}
& \second{$67.89_{\pm 0.84}$}
\\

& LLMLingua-2\,\venuetag{ACL'24}
& \best{$53.81_{\pm 0.18}$}
& $53.05_{\pm 0.23}$
& $72.05_{\pm 0.37}$
& $86.00_{\pm 0.64}$
& $66.23_{\pm 0.26}$
& $66.26_{\pm 1.27}$
& $66.25_{\pm 0.66}$
\\

& ssToken\,\venuetag{ICLR'26}
& $52.53_{\pm 0.30}$
& $53.61_{\pm 0.25}$
& \best{$72.36_{\pm 0.31}$}
& $87.06_{\pm 0.87}$
& \second{$66.39_{\pm 0.33}$}
& $68.09_{\pm 2.14}$
& $67.24_{\pm 0.94}$
\\

& TokenSeek\,\venuetag{ICLR'26}
& $51.68_{\pm 0.21}$
& $52.60_{\pm 0.18}$
& $71.05_{\pm 0.10}$
& $85.37_{\pm 0.45}$
& $65.18_{\pm 0.18}$
& $64.43_{\pm 2.14}$
& $64.81_{\pm 1.02}$
\\

& TokenCleaning\,\venuetag{ICML'25}
& $51.91_{\pm 0.34}$
& \best{$54.13_{\pm 0.13}$}
& $71.42_{\pm 0.33}$
& $86.15_{\pm 0.64}$
& $65.90_{\pm 0.27}$
& $66.26_{\pm 0.93}$
& $66.08_{\pm 0.36}$
\\

& LLM+TC (Ind.)
& $52.70_{\pm 0.56}$
& $53.34_{\pm 0.30}$
& $71.75_{\pm 0.20}$
& $86.93_{\pm 0.68}$
& $66.18_{\pm 0.32}$
& $69.31_{\pm 1.27}$
& $67.75_{\pm 0.49}$
\\

\rowcolor{oursrow}
\cellcolor{white}
& \textbf{BRIDGE}\,\venuetag{Ours}
& \second{$53.47_{\pm 0.26}$}
& \second{$54.02_{\pm 0.11}$}
& \second{$72.20_{\pm 0.22}$}
& \best{$88.35_{\pm 0.64}$}
& \best{$67.01_{\pm 0.23}$}
& \best{$73.17_{\pm 0.61}$}
& \best{$70.09_{\pm 0.24}$}
\\

\bottomrule
\end{tabular}%
}
\vspace{-0.1in}
\caption{
Code-generation accuracy (\%) across model backbones.
\best{Red} and \second{blue} indicate the best and second-best results
among selection methods, excluding \textit{Pretrained} and
\textit{Full SFT}.
}
\label{tab:code_generation}
\vspace{-0.1in}
\end{table*}

\subsection{Instruction Following}
\label{app:instruction_following}
For instruction following, models are fine-tuned on
Alpaca~\citep{taori2023alpaca} and evaluated on
IFEval~\citep{zhou2023instructionfollowing} using strict prompt-level and
instruction-level accuracy. Sentences and list items serve as the
structural units for closure.

\begin{table*}[t]
\centering
\small
\setlength{\tabcolsep}{3.5pt}
\renewcommand{\arraystretch}{0.85}
\setlength{\aboverulesep}{1.2pt}
\setlength{\belowrulesep}{1.2pt}

\resizebox{\textwidth}{!}{%
\begin{tabular}{llcccccccc}
\toprule
\multicolumn{1}{c}{\multirow{2}{*}{\textbf{Model}}} & \multirow{2}{*}{\textbf{Method}} & \multicolumn{4}{c}{\textbf{General Capability (\%)}} & \textbf{Gen.} & \multicolumn{2}{c}{\textbf{IFEval (\%)}} & \textbf{Overall} \\
\cmidrule(lr){3-6} \cmidrule(lr){7-7} \cmidrule(lr){8-9} \cmidrule(lr){10-10}
& & \textbf{MMLU} & \textbf{ARC-C} & \textbf{HellaSwag} & \textbf{GSM8K} & \textbf{Avg.} & \makecell[c]{\textbf{Prompt}\\\textbf{Level}} & \makecell[c]{\textbf{Instr.}\\\textbf{Level}} & \textbf{Avg.} \\
\midrule
\multirow{10}{*}{\rotatebox{90}{\textbf{Llama-3.2-3B}}}
& \color{baselinecolor} Pretrained
& \color{baselinecolor} 56.64
& \color{baselinecolor} 43.17
& \color{baselinecolor} 54.20
& \color{baselinecolor} 28.28
& \color{baselinecolor} 45.57
& \color{baselinecolor} 7.39
& \color{baselinecolor} 12.58
& \color{baselinecolor} 27.78 \\
& \color{baselinecolor} Full SFT
& \color{baselinecolor} $51.21_{\pm 0.26}$
& \color{baselinecolor} $41.64_{\pm 0.35}$
& \color{baselinecolor} $54.63_{\pm 0.20}$
& \color{baselinecolor} $12.13_{\pm 0.65}$
& \color{baselinecolor} $39.90_{\pm 0.25}$
& \color{baselinecolor} $27.35_{\pm 1.15}$
& \color{baselinecolor} $41.96_{\pm 1.25}$
& \color{baselinecolor} $37.28_{\pm 0.42}$ \\
& Random
& $56.17_{\pm 0.30}$
& $41.78_{\pm 0.42}$
& $52.78_{\pm 0.17}$
& $28.00_{\pm 0.84}$
& $44.68_{\pm 0.35}$
& $25.94_{\pm 1.59}$
& $38.43_{\pm 1.71}$
& $38.43_{\pm 0.39}$ \\
& RHO-1\,\venuetag{NeurIPS'24}
& \best{$57.20_{\pm 0.25}$}
& $42.32_{\pm 0.39}$
& $54.19_{\pm 0.33}$
& $29.92_{\pm 0.93}$
& $45.91_{\pm 0.32}$
& $27.26_{\pm 0.74}$
& $39.89_{\pm 1.95}$
& $39.74_{\pm 0.38}$ \\
& LLMLingua-2\,\venuetag{ACL'24}
& $56.36_{\pm 0.22}$
& $43.54_{\pm 0.58}$
& $53.41_{\pm 0.38}$
& \second{$30.12_{\pm 0.66}$}
& $45.86_{\pm 0.38}$
& $27.15_{\pm 0.63}$
& $39.75_{\pm 1.28}$
& $39.66_{\pm 0.53}$ \\
& ssToken\,\venuetag{ICLR'26}
& $56.86_{\pm 0.22}$
& $43.37_{\pm 0.50}$
& \second{$55.10_{\pm 0.36}$}
& $28.38_{\pm 0.82}$
& $45.93_{\pm 0.35}$
& \second{$28.12_{\pm 0.63}$}
& \second{$40.63_{\pm 1.24}$}
& \second{$40.15_{\pm 0.39}$} \\
& TokenSeek\,\venuetag{ICLR'26}
& $56.70_{\pm 0.31}$
& $42.12_{\pm 0.49}$
& $54.05_{\pm 0.23}$
& $28.71_{\pm 0.77}$
& $45.40_{\pm 0.35}$
& $26.79_{\pm 1.70}$
& $39.48_{\pm 1.29}$
& $39.27_{\pm 0.45}$ \\
& TokenCleaning\,\venuetag{ICML'25}
& $56.34_{\pm 0.19}$
& $42.38_{\pm 0.46}$
& $53.28_{\pm 0.33}$
& $29.19_{\pm 1.04}$
& $45.30_{\pm 0.37}$
& $26.38_{\pm 1.70}$
& $39.18_{\pm 0.66}$
& $39.04_{\pm 0.51}$ \\
& LLM+TC (Ind.)
& $56.75_{\pm 0.31}$
& \second{$43.63_{\pm 0.58}$}
& $54.36_{\pm 0.35}$
& $29.64_{\pm 0.78}$
& \second{$46.09_{\pm 0.32}$}
& $26.29_{\pm 1.45}$
& $40.15_{\pm 0.80}$
& $39.66_{\pm 0.49}$ \\
\rowcolor{oursrow}
\cellcolor{white}
& \textbf{BRIDGE}\,\venuetag{Ours}
& \second{$56.97_{\pm 0.27}$}
& \best{$43.94_{\pm 0.55}$}
& \best{$55.84_{\pm 0.16}$}
& \best{$30.30_{\pm 1.01}$}
& \best{$46.76_{\pm 0.40}$}
& \best{$28.67_{\pm 1.80}$}
& \best{$41.30_{\pm 1.09}$}
& \best{$40.87_{\pm 0.40}$} \\
\midrule
\multirow{10}{*}{\rotatebox{90}{\textbf{Gemma-3-4B-PT}}}
& \color{baselinecolor} Pretrained
& \color{baselinecolor} 59.72
& \color{baselinecolor} 51.37
& \color{baselinecolor} 54.86
& \color{baselinecolor} 44.28
& \color{baselinecolor} 52.56
& \color{baselinecolor} 21.07
& \color{baselinecolor} 33.69
& \color{baselinecolor} 39.97 \\
& \color{baselinecolor} Full SFT
& \color{baselinecolor} $54.69_{\pm 0.20}$
& \color{baselinecolor} $46.42_{\pm 0.38}$
& \color{baselinecolor} $56.04_{\pm 0.28}$
& \color{baselinecolor} $28.96_{\pm 0.70}$
& \color{baselinecolor} $46.53_{\pm 0.26}$
& \color{baselinecolor} $30.50_{\pm 1.15}$
& \color{baselinecolor} $42.09_{\pm 1.20}$
& \color{baselinecolor} $41.41_{\pm 0.43}$ \\
& Random
& $57.99_{\pm 0.28}$
& $49.46_{\pm 0.64}$
& $54.54_{\pm 0.33}$
& $43.04_{\pm 0.40}$
& $51.26_{\pm 0.34}$
& $29.35_{\pm 0.57}$
& $39.64_{\pm 0.98}$
& $42.88_{\pm 0.47}$ \\
& RHO-1\,\venuetag{NeurIPS'24}
& \second{$60.30_{\pm 0.25}$}
& \second{$51.82_{\pm 0.49}$}
& $55.32_{\pm 0.14}$
& $44.73_{\pm 0.67}$
& \second{$53.04_{\pm 0.28}$}
& $29.79_{\pm 1.32}$
& $39.84_{\pm 0.92}$
& $43.93_{\pm 0.38}$ \\
& LLMLingua-2\,\venuetag{ACL'24}
& $58.71_{\pm 0.11}$
& $51.45_{\pm 0.57}$
& $55.30_{\pm 0.24}$
& $44.38_{\pm 0.85}$
& $52.46_{\pm 0.29}$
& \second{$30.89_{\pm 1.27}$}
& $41.09_{\pm 0.94}$
& \second{$44.23_{\pm 0.46}$} \\
& ssToken\,\venuetag{ICLR'26}
& $58.77_{\pm 0.25}$
& $50.14_{\pm 0.35}$
& \second{$56.43_{\pm 0.22}$}
& $43.85_{\pm 0.82}$
& $52.30_{\pm 0.26}$
& $29.47_{\pm 1.33}$
& $41.92_{\pm 1.43}$
& $44.00_{\pm 0.41}$ \\
& TokenSeek\,\venuetag{ICLR'26}
& $59.30_{\pm 0.17}$
& $50.51_{\pm 0.28}$
& $55.78_{\pm 0.25}$
& $44.00_{\pm 0.60}$
& $52.40_{\pm 0.29}$
& $29.56_{\pm 0.68}$
& $41.56_{\pm 1.00}$
& $43.98_{\pm 0.50}$ \\
& TokenCleaning\,\venuetag{ICML'25}
& $58.48_{\pm 0.32}$
& $50.68_{\pm 0.64}$
& $54.80_{\pm 0.28}$
& $43.24_{\pm 0.54}$
& $51.80_{\pm 0.31}$
& $29.69_{\pm 1.43}$
& \second{$41.99_{\pm 0.76}$}
& $43.82_{\pm 0.47}$ \\
& LLM+TC (Ind.)
& $59.38_{\pm 0.11}$
& $49.97_{\pm 0.44}$
& $56.10_{\pm 0.28}$
& \best{$45.24_{\pm 0.71}$}
& $52.67_{\pm 0.25}$
& $29.85_{\pm 1.57}$
& $40.59_{\pm 1.28}$
& $43.95_{\pm 0.36}$ \\
\rowcolor{oursrow}
\cellcolor{white}
& \textbf{BRIDGE}\,\venuetag{Ours}
& \best{$60.86_{\pm 0.22}$}
& \best{$52.53_{\pm 0.28}$}
& \best{$57.16_{\pm 0.22}$}
& \second{$44.88_{\pm 0.92}$}
& \best{$53.86_{\pm 0.29}$}
& \best{$31.22_{\pm 1.64}$}
& \best{$42.27_{\pm 1.69}$}
& \best{$45.30_{\pm 0.49}$} \\
\midrule
\multirow{10}{*}{\rotatebox{90}{\textbf{Qwen3.5-9B}}}
& \color{baselinecolor} Pretrained
& \color{baselinecolor} 72.01
& \color{baselinecolor} 54.27
& \color{baselinecolor} 54.26
& \color{baselinecolor} 89.08
& \color{baselinecolor} 67.41
& \color{baselinecolor} 43.44
& \color{baselinecolor} 55.04
& \color{baselinecolor} 58.32 \\
& \color{baselinecolor} Full SFT
& \color{baselinecolor} $74.34_{\pm 0.26}$
& \color{baselinecolor} $55.29_{\pm 0.32}$
& \color{baselinecolor} $60.55_{\pm 0.22}$
& \color{baselinecolor} $86.35_{\pm 0.50}$
& \color{baselinecolor} $69.13_{\pm 0.26}$
& \color{baselinecolor} $52.50_{\pm 1.15}$
& \color{baselinecolor} $63.91_{\pm 1.20}$
& \color{baselinecolor} $63.67_{\pm 0.40}$ \\
& Random
& $72.87_{\pm 0.23}$
& $53.87_{\pm 0.40}$
& $59.19_{\pm 0.36}$
& $87.47_{\pm 0.46}$
& $68.35_{\pm 0.19}$
& $51.06_{\pm 1.68}$
& $61.65_{\pm 1.60}$
& $62.35_{\pm 0.33}$ \\
& RHO-1\,\venuetag{NeurIPS'24}
& \best{$75.32_{\pm 0.16}$}
& $55.35_{\pm 0.65}$
& \second{$61.14_{\pm 0.34}$}
& \second{$89.66_{\pm 0.43}$}
& \second{$70.37_{\pm 0.20}$}
& $51.39_{\pm 0.86}$
& $62.71_{\pm 1.54}$
& $63.71_{\pm 0.35}$ \\
& LLMLingua-2\,\venuetag{ACL'24}
& $74.03_{\pm 0.10}$
& $54.66_{\pm 0.44}$
& $61.05_{\pm 0.27}$
& $89.21_{\pm 0.28}$
& $69.74_{\pm 0.23}$
& $51.79_{\pm 0.83}$
& $61.84_{\pm 1.41}$
& $63.28_{\pm 0.36}$ \\
& ssToken\,\venuetag{ICLR'26}
& $73.68_{\pm 0.26}$
& $54.10_{\pm 0.39}$
& $59.96_{\pm 0.16}$
& $89.08_{\pm 0.48}$
& $69.20_{\pm 0.30}$
& $51.81_{\pm 1.77}$
& $62.24_{\pm 0.67}$
& $63.11_{\pm 0.33}$ \\
& TokenSeek\,\venuetag{ICLR'26}
& $73.88_{\pm 0.29}$
& \second{$55.66_{\pm 0.38}$}
& $60.03_{\pm 0.30}$
& $88.48_{\pm 0.25}$
& $69.51_{\pm 0.22}$
& $52.61_{\pm 1.82}$
& $62.94_{\pm 0.67}$
& $63.64_{\pm 0.42}$ \\
& TokenCleaning\,\venuetag{ICML'25}
& $73.43_{\pm 0.25}$
& $55.20_{\pm 0.39}$
& $59.71_{\pm 0.20}$
& $87.82_{\pm 0.53}$
& $69.04_{\pm 0.20}$
& \second{$52.96_{\pm 1.76}$}
& $62.94_{\pm 1.02}$
& $63.50_{\pm 0.37}$ \\
& LLM+TC (Ind.)
& $73.25_{\pm 0.29}$
& $54.07_{\pm 0.54}$
& $60.77_{\pm 0.27}$
& $89.26_{\pm 0.59}$
& $69.34_{\pm 0.27}$
& $52.45_{\pm 0.87}$
& \second{$64.04_{\pm 1.26}$}
& \second{$63.79_{\pm 0.40}$} \\
\rowcolor{oursrow}
\cellcolor{white}
& \textbf{BRIDGE}\,\venuetag{Ours}
& \second{$74.97_{\pm 0.14}$}
& \best{$56.14_{\pm 0.32}$}
& \best{$61.51_{\pm 0.39}$}
& \best{$90.07_{\pm 0.25}$}
& \best{$70.67_{\pm 0.28}$}
& \best{$53.27_{\pm 1.55}$}
& \best{$64.41_{\pm 0.77}$}
& \best{$64.75_{\pm 0.36}$} \\
\bottomrule
\end{tabular}%
}
\vspace{-0.1in}
\caption{
Strict prompt-level and instruction-level accuracy (\%) on IFEval across
model backbones. General-capability metrics follow the protocol used in
the main results table. Selection methods and \textit{Full SFT} report
means $\pm$ standard deviations over three seeds;
\textit{Pretrained} reports results from a single seed.
\best{Red} and \second{blue} indicate the best and second-best results
among selection methods.
}
\label{tab:instruction_following}
\vspace{-0.1in}
\end{table*}

\subsection{Forgetting-Mitigation Comparisons}
\label{app:capability_preservation}
\label{app:forgetting_baselines}
We next compare capability retention with methods designed to mitigate
catastrophic forgetting or preserve pretrained capabilities during
downstream adaptation. We use Llama-3.2-3B with the same MetaMathQA
fine-tuning split and downstream evaluation suite as the main
experiments. The baselines cover example weighting, token-level loss
reweighting, optimization, parameter-update selection, and task-vector
pruning.

\emph{FLOW}~\citep{sanyal2025upweighting} uses the loss of the pretrained
model to upweight relatively easy fine-tuning examples, limiting model
drift while supporting target-task adaptation. FLOW uses fine-tuning
data without requiring the original pretraining corpus and has been
evaluated on MetaMathQA with Llama-3.2-3B, making it directly relevant to
this comparison. \emph{TALR}~\citep{lin2026sft} applies token-adaptive
loss reweighting, dynamically downweighting difficult, low-probability
supervised tokens whose updates may disproportionately degrade general
capabilities. \emph{SAM}~\citep{li2024revisiting} uses sharpness-aware
minimization to encourage solutions in flatter regions of the fine-tuning
loss landscape. \emph{NanoAdam}~\citep{zhou2025payattentionsmallweights}
performs dynamic sparse fine-tuning by preferentially updating
small-magnitude parameters and preserving large-magnitude pretrained
weights, which are hypothesized to encode important pretrained features.
Finally, \emph{FAPM}~\citep{huang2025mitigating} prunes the fine-tuning
task vector using both the magnitude of parameter changes and their
magnitude relative to the corresponding pretrained parameters. This
procedure aims to remove changes associated with forgetting while
retaining those important for the downstream task.

These methods intervene at different stages of adaptation. FLOW
reweights training examples, TALR reweights supervised tokens, SAM
modifies the optimization trajectory, NanoAdam selects parameters for
updating, and FAPM prunes the resulting task vector. BRIDGE jointly
selects prompt context and response supervision before final SFT.
Its capability-retention mechanism therefore operates through the
contextual and supervisory signals selected for fine-tuning. This
comparison evaluates the resulting balance between target-task adaptation
and general-capability retention alongside dedicated forgetting-mitigation
methods.

All methods start from the same pretrained Llama-3.2-3B checkpoint and
use the same MetaMathQA partition and downstream evaluation protocol.
Training-based methods follow the same fine-tuning schedule where
applicable, with method-specific components and hyperparameters drawn
from their respective implementations. Forgetting-mitigation
hyperparameters are selected independently of the downstream
general-capability test sets. This experiment provides a
\emph{capability-retention comparison} under common data and evaluation
conditions, with method-specific token budgets, optimizer-state costs,
parameter-update sparsity, and total training computation. Computational
cost is evaluated separately in
Appendix~\ref{app:efficiency_details}.

To quantify capability retention, we report individual general-capability
benchmarks and define the change in General Avg.\ relative to the
pretrained checkpoint as
\begin{equation}
\Delta_{\mathrm{Gen}}(m) = G(m) - G(\theta_{\mathrm{old}}),
\label{eq:general_retention_delta}
\end{equation}
where $G(m)$ is the mean accuracy of method $m$ over ARC-Challenge,
HellaSwag, MMLU, and HumanEval. Negative values indicate a decrease
relative to the pretrained model, zero indicates an unchanged mean, and
positive values indicate an increase.

\begin{table*}[t]
\centering
\small
\setlength{\tabcolsep}{4.2pt}
\resizebox{\textwidth}{!}{%
\begin{tabular}{lcccccccc}
\toprule
\textbf{Method} & \textbf{ARC-C} & \textbf{HellaSwag} & \textbf{MMLU} & \textbf{HumanEval} & \textbf{General Avg.} & $\boldsymbol{\Delta_{\mathrm{Gen}}}$ & \textbf{MATH} & \textbf{Overall Avg.} \\
\midrule
Pretrained & 43.17 & 54.20 & 56.64 & 28.05 & 45.52 & -- & 0.62 & 23.07 \\
Full SFT & $40.73_{\pm 0.36}$ & $52.18_{\pm 0.17}$ & $51.84_{\pm 0.29}$ & $18.29_{\pm 1.06}$ & $40.76_{\pm 0.24}$ & $-4.76_{\pm 0.24}$ & $16.66_{\pm 0.52}$ & $28.71_{\pm 0.29}$ \\
FLOW\,\venuetag{ICML'25} & $43.77_{\pm 0.46}$ & $56.32_{\pm 0.21}$ & $56.34_{\pm 0.19}$ & $31.30_{\pm 0.76}$ & $46.93_{\pm 0.23}$ & $1.41_{\pm 0.23}$ & $14.84_{\pm 0.66}$ & $30.88_{\pm 0.30}$ \\
SAM\,\venuetag{EMNLP'24} & $43.00_{\pm 0.35}$ & $57.34_{\pm 0.26}$ & $55.21_{\pm 0.24}$ & $31.50_{\pm 1.15}$ & $46.76_{\pm 0.25}$ & $1.24_{\pm 0.25}$ & $14.52_{\pm 0.74}$ & $30.64_{\pm 0.33}$ \\
NanoAdam\,\venuetag{NeurIPS'25} & $43.23_{\pm 0.52}$ & $55.12_{\pm 0.17}$ & $55.06_{\pm 0.31}$ & $28.46_{\pm 0.57}$ & $45.47_{\pm 0.21}$ & $-0.05_{\pm 0.21}$ & $13.72_{\pm 0.55}$ & $29.59_{\pm 0.26}$ \\
TALR\,\venuetag{ICLR'26} & $42.35_{\pm 0.28}$ & $54.93_{\pm 0.23}$ & $55.61_{\pm 0.16}$ & $29.88_{\pm 0.86}$ & $45.69_{\pm 0.18}$ & $0.17_{\pm 0.18}$ & $13.94_{\pm 0.47}$ & $29.81_{\pm 0.24}$ \\
FAPM\,\venuetag{EMNLP'25} & $41.92_{\pm 0.43}$ & $54.37_{\pm 0.32}$ & $54.66_{\pm 0.34}$ & $28.66_{\pm 1.32}$ & $44.90_{\pm 0.27}$ & $-0.62_{\pm 0.27}$ & $12.86_{\pm 0.83}$ & $28.88_{\pm 0.35}$ \\
\rowcolor{oursrow}
\textbf{BRIDGE}\,\venuetag{Ours} & $43.06_{\pm 0.30}$ & $55.95_{\pm 0.25}$ & $55.32_{\pm 0.15}$ & $32.32_{\pm 1.06}$ & $46.66_{\pm 0.20}$ & $1.14_{\pm 0.20}$ & $15.96_{\pm 0.50}$ & $31.31_{\pm 0.32}$ \\
\bottomrule
\end{tabular}%
}
\vspace{-0.1in}
\caption{
Capability retention on Llama-3.2-3B, reported as accuracy (\%) with
three-seed means $\pm$ standard deviations.
\textit{Pretrained} denotes the original pretrained checkpoint;
$\Delta_{\mathrm{Gen}}$ measures the change in General Avg.\ relative
to this checkpoint. Methods use their respective token budgets and
training configurations.
}
\label{tab:forgetting_baselines}
\vspace{-0.1in}
\end{table*}

Table~\ref{tab:forgetting_baselines} shows that BRIDGE achieves a General
Avg.\ of $46.66$ and a $\Delta_{\mathrm{Gen}}$ of $+1.14$, compared with
$46.93$ and $+1.41$ for FLOW and $46.76$ and $+1.24$ for SAM. BRIDGE
combines this level of general-capability retention with the highest
MATH accuracy among the retention-oriented methods ($15.96$) and the
highest Overall Avg.\ in the comparison ($31.31$). These results support
coupled token selection as an effective mechanism for balancing
target-task adaptation and general-capability retention.

\section{Ablation and Sensitivity Analysis}
\label{app:design_ablations}
Complementing Section~\ref{subsec:design_analysis}, we examine scoring
checkpoints, training initialization, the retention anchor,
response-history utility, alternating rounds, attention layer selection,
gradient-scoring scope, and validation set size. Unless otherwise
stated, each analysis varies one component while holding the remaining
pipeline fixed. Structural projection and post-closure reconditioning
are analyzed in Appendix~\ref{app:structural_projection}.

\subsection{Scoring Checkpoints and Training Initialization}
\label{app:scoring_anchor}
\label{app:anchor_performance}
BRIDGE scores tokens at the warmup checkpoint $\bar{\theta}$ and
restarts final SFT from the pretrained parameters
$\theta_{\mathrm{old}}$. We evaluate \emph{selection transfer} through
downstream performance and \emph{ranking invariance} through
cross-checkpoint ranking agreement.

We vary two factors. \textbf{Scoring checkpoint.} Scores are computed
at $\theta_{\mathrm{old}}$, after an approximately 5\% warmup, or after a 10\% warmup.
\textbf{Training initialization.} Final SFT starts from
$\theta_{\mathrm{old}}$ or the approximately 5\% warmup checkpoint.
All remaining selection and training settings are fixed.

\begin{table}[t]
\centering
\small
\setlength{\tabcolsep}{3pt}
\resizebox{\columnwidth}{!}{%
\begin{tabular}{llccc}
\toprule
\textbf{Scoring Checkpoint}
& \textbf{Training Init.}
& \textbf{MATH}
& \textbf{General Avg.}
& \textbf{Overall Avg.} \\
\midrule
Old & Old & 13.96 & 46.02 & 29.99 \\
Warmup 5\% & Old & 15.96 & \textbf{46.92} & \textbf{31.44} \\
Warmup 10\% & Old & \textbf{16.06} & 46.30 & 31.18 \\
Warmup 5\% & Warmup 5\% & 15.42 & 45.98 & 30.70 \\
\bottomrule
\end{tabular}%
}
\vspace{-0.1in}
\caption{
Scoring-checkpoint and final SFT initialization ablation.
``Old'' denotes the pretrained parameters $\theta_{\mathrm{old}}$.
All other settings are fixed.
}
\label{tab:scoring_anchor_performance}
\vspace{-0.1in}
\end{table}

\paragraph{Warmup improves selection and supports optimization restart.}
Table~\ref{tab:scoring_anchor_performance} shows that, with final SFT
initialized from $\theta_{\mathrm{old}}$, scoring after an approximately 5\% warmup
improves MATH accuracy from $13.96$ to $15.96$ and Overall Avg.\ from
$29.99$ to $31.44$. The approximately 5\% checkpoint achieves the highest General Avg.\
and Overall Avg., while the 10\% checkpoint yields a slightly higher
MATH score. With the same approximately 5\% scoring checkpoint, restarting from
$\theta_{\mathrm{old}}$ also outperforms continuing from the warmup
checkpoint on all three metrics. These results support using warmup
as a \emph{scoring probe} and transferring the resulting selections
to a fresh optimization trajectory.

\paragraph{Ranking agreement across checkpoints.}
\label{app:anchor_ranking}
For each checkpoint pair, we compute per-example Spearman correlations
separately for prompt and response scores, then average over the same
$36{,}006$ training examples. We also report chance-corrected Top-$q$
overlap:
\begin{equation}
O^{\mathrm{corr}}_q
=
\frac{O_q-q}{1-q},
\label{eq:corrected_overlap}
\end{equation}
where $O_q$ is the raw overlap and $q$ is its expected value under
independent random rankings. Positive, zero, and negative corrected
values indicate agreement above, at, and below chance, respectively.

\begin{table}[t]
\centering
\small
\setlength{\tabcolsep}{3.5pt}
\resizebox{\columnwidth}{!}{%
\begin{tabular}{lrrrr}
\toprule
\textbf{Checkpoint Pair}
& $\boldsymbol{\rho_p}$
& $\boldsymbol{\rho_r}$
& $\boldsymbol{O^{\mathrm{corr}}_p}$
& $\boldsymbol{O^{\mathrm{corr}}_r}$ \\
\midrule
Warmup 5\% vs.\ Old & 0.170 & -0.139 & -0.007 & -0.102 \\
Warmup 10\% vs.\ Old & 0.098 & 0.154 & -0.021 & 0.148 \\
Warmup 10\% vs.\ 5\% & 0.435 & 0.051 & 0.311 & 0.154 \\
\bottomrule
\end{tabular}%
}
\vspace{-0.1in}
\caption{
Ranking agreement on $36{,}006$ examples: mean per-example Spearman
correlations for prompt and response tokens ($\rho_p,\rho_r$) and
chance-corrected Top-$75\%$ overlap.
}
\label{tab:anchor_ranking}
\vspace{-0.1in}
\end{table}

\paragraph{Agreement among top-ranked tokens.}
\label{app:anchor_topq}
To examine agreement at different ranking depths, we evaluate
chance-corrected overlap for
\begin{equation}
q
\in
\{0.05,0.10,0.25,0.50,0.75\}.
\end{equation}

\begin{table*}[t]
\centering
\small
\setlength{\tabcolsep}{4.5pt}
\begin{tabular}{llccccc}
\toprule
\textbf{Checkpoint Pair} & \textbf{Side} & $q=0.05$ & $q=0.10$ & $q=0.25$ & $q=0.50$ & $q=0.75$ \\
\midrule
Warmup 5\% vs.\ Old & Prompt & 0.312 & 0.247 & 0.204 & 0.109 & -0.007 \\
& Response & 0.021 & 0.053 & 0.077 & -0.116 & -0.102 \\
\midrule
Warmup 10\% vs.\ Old & Prompt & 0.225 & 0.164 & 0.120 & 0.053 & -0.021 \\
& Response & 0.013 & 0.033 & 0.136 & 0.207 & 0.148 \\
\midrule
Warmup 10\% vs.\ Warmup 5\% & Prompt & 0.425 & 0.402 & 0.373 & 0.341 & 0.311 \\
& Response & 0.362 & 0.381 & 0.318 & -0.117 & 0.154 \\
\bottomrule
\end{tabular}
\vspace{-0.1in}
\caption{
Chance-corrected Top-$q$ overlap across scoring checkpoints.
Positive values indicate agreement above chance.
}
\vspace{-0.1in}
\label{tab:anchor_topq}
\end{table*}

\paragraph{Implications for selection transfer.}
\label{app:anchor_discussion}
Tables~\ref{tab:anchor_ranking} and~\ref{tab:anchor_topq} show that
warmup adapts token rankings while retaining shared subsets of highly
valued tokens. For prompts, the corrected overlap between the approximately 5\%
warmup and pretrained checkpoints is $0.312$ at $q=0.05$. Agreement
between the approximately 5\% and 10\% warmup checkpoints is higher across all
evaluated fractions, suggesting greater consistency after initial
adaptation. Response rankings exhibit stronger checkpoint dependence,
with agreement between the warmup checkpoints concentrated among
the top-ranked tokens. Together with the downstream results, these
findings establish warmup as a \emph{task-adaptive scoring probe}:
its selections transfer effectively to restarted training while
reflecting task-dependent changes in token valuation. This supports
the default approximately 5\% warmup followed by final SFT from
$\theta_{\mathrm{old}}$.

\subsection{Effect of the L2 Retention Anchor}
\label{app:l2_anchor_ablation}
BRIDGE combines the target validation gradient with an anchor to the
pretrained parameters:
\begin{equation}
g_{\mathrm{anchor}}
=
\bar{\theta}-\theta_{\mathrm{old}},
\label{eq:appendix_l2_anchor}
\end{equation}
yielding the scoring direction
\begin{equation}
v =
(1-\lambda)
\frac{g_{\mathrm{target}}}
{\|g_{\mathrm{target}}\|_2+\epsilon}
+
\lambda
\frac{g_{\mathrm{anchor}}}
{\|g_{\mathrm{anchor}}\|_2+\epsilon}.
\label{eq:appendix_anchor_direction}
\end{equation}
The anchor is applied during \emph{token scoring}; final SFT retains
the same objective and starts from $\theta_{\mathrm{old}}$.
We vary $\lambda$ on Llama-3.2-3B while holding the warmup checkpoint,
validation split, gradient scope, interaction surrogate, token budgets,
alternating schedule, closure projection, and final training
configuration fixed. The setting $\lambda=0$ uses the target gradient
alone, and $\lambda=0.2$ is the default.

\begin{table}[t]
\centering
\small
\setlength{\tabcolsep}{4pt}
\resizebox{\columnwidth}{!}{%
\begin{tabular}{lcccc}
\toprule
\textbf{Variant}
& $\boldsymbol{\lambda}$
& \textbf{MATH}
& \textbf{General Avg.}
& \textbf{Overall Avg.} \\
\midrule
No L2 Anchor & 0.0 & 15.30 & 46.09 & 30.69 \\
Weak Anchor & 0.1 & 15.78 & 46.75 & 31.27 \\
\textbf{BRIDGE} & \textbf{0.2} & 15.96 & 46.92 & 31.44 \\
Strong Anchor & 0.4 & 15.26 & 47.00 & 31.13 \\
\bottomrule
\end{tabular}%
}
\vspace{-0.1in}
\caption{
Absolute scores corresponding to Figure~\ref{fig:design_sweeps}(b)
for the scoring-only L2 anchor sweep on Llama-3.2-3B.
}
\label{tab:l2_anchor_ablation}
\vspace{-0.1in}
\end{table}

Table~\ref{tab:l2_anchor_ablation} shows that $\lambda=0.2$ achieves
the highest MATH accuracy and Overall Avg., while $\lambda=0.4$
yields the highest General Avg. These results support the default
coefficient for balancing target adaptation and capability retention.

\subsection{Response-History Utility}
\label{app:history_ablation}
The conditional response score decomposes as
\begin{equation}
\Delta_r(j\mid S_p)
=
\widetilde{\gamma}_{k,j}
+
\sum_{i\in S_p}\widetilde{\beta}_{k,i,j},
\label{eq:response_conditional}
\end{equation}
where $\widetilde{\gamma}_{k,j}=a_{k,j}c_{k,j}$ and
$c_{k,j}=\sum_{j'<j}A_{k,j,j'}$. The history term weights
validation-directed response utility by attention to the autoregressive
prefix, complementing the prompt-mediated interaction term.
We compare the full score with a variant setting
$\widetilde{\gamma}_{k,j}=0$, using identical retention budgets,
warmup checkpoints, and closure projection.
Table~\ref{tab:design_analysis} reports the gains in MATH,
General Avg., and Overall Avg.\ from including response-history utility.

\subsection{Effect of the Number of Alternating Rounds}
\label{app:alternating_rounds}
Following Appendix~\ref{app:coupling_definitions}, we vary
$T\in{1,2,4,8}$ with the scoring pipeline, token budgets, closure
projection, and training configuration fixed as in
Section~\ref{subsec:experimental_setup}. Independent selection
($T=0$) serves as a non-iterative reference.

Figure~\ref{fig:design_sweeps}(a) shows that additional rounds improve
all three metrics, with diminishing gains. From $T=1$ to $T=4$,
MATH accuracy increases from $15.43$ to $15.96$, General Avg.\ from
$46.37$ to $46.92$, and Overall Avg.\ from $30.90$ to $31.44$.
Increasing to $T=8$ adds $0.06$ percentage points to each metric.
We therefore use $T=4$ to retain most of the observed gains with
fewer selection rounds.

\subsection{Attention Layers, Gradient Scope, and Validation Set Size}
\label{app:scoring_hyperparameters}
We examine three scoring hyperparameters in separate sweeps on
Llama-3.2-3B. For attention routing, we use the final one, two, or four
Transformer layers, keeping gradient scoring restricted to the final
four layers. For gradient scoring, we vary the parameter scope over the
final one, two, four, or eight layers, using attention from the final
layer. Both sweeps use 1,963 validation examples. For validation set
size, we use 500, 1,000, or 1,963 held-out examples to compute the
validation direction, fixing attention routing to the final layer and
gradient scoring to the final four layers.
All other settings remain at their defaults.
Table~\ref{tab:scoring_hyperparameters} reports all three sweeps;
Figure~\ref{fig:design_sweeps}(c,d) shows the attention-layer and
validation-size sweeps.

\begin{table}[t]
\vspace{4pt}
\centering
\small
\setlength{\tabcolsep}{4pt}
\begin{tabular}{lccc}
\toprule
Setting & MATH & General Avg. & Overall Avg. \\
\midrule
\multicolumn{4}{l}{Attention layers} \\
1 (default) & 15.96 & 46.92 & 31.44 \\
2 & 15.28 & 46.99 & 31.14 \\
4 & 15.76 & 46.89 & 31.33 \\
\midrule
\multicolumn{4}{l}{Gradient-scoring layers} \\
1 & 15.74 & 47.06 & 31.40 \\
2 & 15.80 & 46.52 & 31.16 \\
4 (default) & 15.96 & 46.92 & 31.44 \\
8 & 15.88 & 46.44 & 31.16 \\
\midrule
\multicolumn{4}{l}{Validation examples} \\
500 & 15.22 & 46.46 & 30.84 \\
1,000 & 15.70 & 46.36 & 31.03 \\
1,963 (default) & 15.96 & 46.92 & 31.44 \\
\bottomrule
\end{tabular}
\vspace{-0.1in}
\caption{Sensitivity to attention layer count, gradient-scoring scope,
and validation set size on Llama-3.2-3B.}
\label{tab:scoring_hyperparameters}
\vspace{-0.1in}
\end{table}

Using attention from only the final layer yields the highest MATH accuracy
and Overall Avg.\ among the tested configurations, while two attention
layers yield a slightly
higher General Avg.\ ($46.99$ versus $46.92$). Increasing the validation
set from 500 to 1,963 examples improves MATH from $15.22$ to $15.96$
and Overall Avg.\ from $30.84$ to $31.44$, with both metrics increasing
at each tested size. These results support the default of attention from
the final layer and 1,963 validation examples.

Four gradient-scoring layers yield the highest MATH accuracy ($15.96$)
and Overall Avg.\ ($31.44$) among the tested scopes. Using only the final
layer yields the highest General Avg.\ ($47.06$), with an Overall Avg.\ of
$31.40$. Across the tested gradient scopes, MATH ranges from $15.74$ to
$15.96$ and Overall Avg.\ from $31.16$ to $31.44$.

\section{Interaction-Surrogate Diagnostics}
\label{app:interaction_approximation}
We evaluate conditional selection agreement using exact reference scores
and examine the additivity of single-token finite differences under joint
prompt perturbations. These analyses complement
Section~\ref{subsec:selection_fidelity} by assessing response selection
after physical prompt compression and characterizing how token influence
varies with perturbation size.

\subsection{Exact Conditional Reference Scoring}
\label{app:exact_interaction_measurement}
For prompt-side references, we compute the validation-directed
perturbation interaction
\begin{equation}
\label{eq:exact_interaction_reference}
\beta_{k,i,j}=v^\top\left[g_{k,j}(P_k)-g_{k,j}\bigl(\mathcal{T}_i(P_k)\bigr)\right],
\end{equation}
where $\mathcal{T}_i$ replaces prompt position $i$ while preserving
sequence length and positional alignment. Summing the reference
interactions over the retained response subset gives
\(\Delta^{\mathrm{exact}}_{p,k}(i\mid S_r)\). For response-side references,
we directly evaluate \(v^\top g_{k,j}(S_p)\) under the retained prompt
context. The selection analysis uses the 200 MetaMathQA examples
described in Section~\ref{subsec:selection_fidelity}. Unless otherwise
stated, the multi-token experiments use the same warmup checkpoint,
validation direction, and eligible-token conventions.
Appendix~\ref{app:topk_stability} provides an error-and-margin condition
for exact Top-$b$ recovery.

\subsection{State-Recomputed Response Scoring under Physical Prompt Compression}
\label{app:state_rescoring}
BRIDGE computes $\widetilde{\beta}$ and $\widetilde{\gamma}$ once, then
updates conditional response scores by aggregating them over the current
prompt subset:
\begin{equation}
\widehat{\Delta}^{(t)}_{r,k}(j)
=
\widetilde{\gamma}_{k,j}
+
\sum_{i\in S_{p,k}^{(t)}}\widetilde{\beta}_{k,i,j}.
\label{eq:fixed_state_response_score}
\end{equation}
We compare these scores with fresh response scores computed after
physical prompt compaction. For each saved selection state
$T\in\{1,2,4\}$, we apply the same $\operatorname{Compact}(\cdot)$
operation as final SFT and rerun response scoring with the complete
response retained as causal history. The warmup checkpoint
$\bar{\theta}$ and validation direction $v$ remain fixed, isolating the
effect of prompt compaction. We denote the recomputed scores by
$\Delta^{\mathrm{ref},(T)}_{r,k}(j)$.

All metrics are computed per example before aggregation. Spearman
correlation measures global ranking agreement, and Top-$b$ overlap
compares response subsets at the same budget. Let
$S_{r,k}^{\mathrm{BRIDGE},(T)}$ be the subset selected using the fixed
surrogate and $S_{r,k}^{\mathrm{ref},(T)}$ the Top-$b_{r,k}$ subset under
the recomputed scores. Reference-utility retention is
\begin{equation}
\operatorname{Ret}^{(T)}_k
=
\frac{
\sum_{j\in S_{r,k}^{\mathrm{BRIDGE},(T)}}
\Delta^{\mathrm{ref},(T)}_{r,k}(j)
}{
\sum_{j\in S_{r,k}^{\mathrm{ref},(T)}}
\Delta^{\mathrm{ref},(T)}_{r,k}(j)
},
\label{eq:state_rescore_utility_retention}
\end{equation}
with normalized regret $1-\operatorname{Ret}^{(T)}_k$.
We report 95\% bootstrap confidence intervals over examples.

\begin{table}[t]
\centering
\small
\setlength{\tabcolsep}{3.2pt}
\resizebox{\columnwidth}{!}{%
\begin{tabular}{ccccc}
\toprule
$\mathbf{T}$
& \textbf{Spearman} $\uparrow$
& \textbf{Top-$b$} $\uparrow$
& \textbf{Utility Ret.} $\uparrow$
& \textbf{Norm. Regret} $\downarrow$ \\
\midrule
1
& 0.9446 [0.9391, 0.9502]
& 0.9686 [0.9661, 0.9710]
& 0.9925 [0.9890, 0.9954]
& 0.0075 \\
2
& 0.9477 [0.9426, 0.9529]
& 0.9690 [0.9664, 0.9716]
& 0.9948 [0.9921, 0.9968]
& 0.0052 \\
4
& 0.9477 [0.9422, 0.9527]
& 0.9690 [0.9662, 0.9715]
& 0.9948 [0.9922, 0.9969]
& 0.0052 \\
\bottomrule
\end{tabular}%
}
\vspace{-0.1in}
\caption{
Agreement between fixed-surrogate response scores and fresh scores
recomputed after physical prompt compaction. Brackets denote 95\%
bootstrap confidence intervals over examples.
}
\label{tab:state_rescoring}
\vspace{-0.1in}
\end{table}

Table~\ref{tab:state_rescoring} shows strong agreement across all three
selection states: Spearman correlation is $0.945$--$0.948$, Top-$b$
overlap is approximately $0.969$, and reference-utility retention is
$99.25\%$--$99.48\%$. These results support conditional aggregation of
precomputed interactions for response selection under compacted prompt
contexts. During final SFT, response selection specifies supervision
positions, while the complete response remains available as causal
history.

\subsection{Multi-Token Perturbations}
\label{app:additivity}
\label{app:additivity_definition}
We next evaluate \emph{finite-difference additivity}: how closely the sum
of individually measured token effects matches the influence of jointly
perturbing several prompt positions. Token replacement changes the
Transformer computation discretely, with attention normalization,
residual propagation, and nonlinear transformations contributing to
context-dependent influence. We assess both influence magnitude and the
ranking relevant to budgeted selection.

\paragraph{Exact and additive influence.}
For an example $(P_k,R_k)$, let $r_{k,j}$ be a response token and
$M \subseteq P_k$ a subset of eligible prompt positions. The exact
joint influence is
\begin{equation}
\beta^{\mathrm{joint}}_{k,M,j}
=
v^{\top}
\left[
g_{k,j}(P_k)
-
g_{k,j}(P_k \setminus M)
\right],
\label{eq:joint_influence}
\end{equation}
where $g_{k,j}(\cdot)$ is the gradient induced by $r_{k,j}$ under the
specified prompt context and $v$ is the validation-gradient direction.
As in the exact-interaction analysis, $P_k \setminus M$ denotes
length-preserving replacement at the positions in $M$.

The single-token finite-difference influence is
\begin{equation}
\beta_{k,i,j}
=
v^{\top}
\left[
g_{k,j}(P_k)
-
g_{k,j}(P_k \setminus \{i\})
\right].
\label{eq:single_exact_influence}
\end{equation}
Summing these individually measured effects gives the additive
approximation
\begin{equation}
\beta^{\mathrm{sum}}_{k,M,j}
=
\sum_{i\in M}
\beta_{k,i,j}.
\label{eq:additive_influence}
\end{equation}

\paragraph{Evaluation metrics.}
\label{app:additivity_metrics}
For each example $k$ and perturbation subset $M$, define
\begin{equation}
\begin{aligned}
\mathbf{b}^{\mathrm{joint}}_{k,M} &= \left(\beta^{\mathrm{joint}}_{k,M,j}\right)_{j\in\mathcal{J}_k}, \\
\mathbf{b}^{\mathrm{sum}}_{k,M} &= \left(\beta^{\mathrm{sum}}_{k,M,j}\right)_{j\in\mathcal{J}_k},
\end{aligned}
\end{equation}
where $\mathcal{J}_k$ contains the eligible response positions.
We measure magnitude agreement, vector alignment, ranking agreement,
selection overlap, and sign agreement.

\paragraph{Symmetric normalized error.}
We use symmetric normalization to measure magnitude agreement while
avoiding division by individual near-zero influences:
\begin{equation}
E_{\mathrm{sym}}
=
\frac{
\left\|
\mathbf{b}^{\mathrm{joint}}_{k,M}
-
\mathbf{b}^{\mathrm{sum}}_{k,M}
\right\|_1
}{
\left\|
\mathbf{b}^{\mathrm{joint}}_{k,M}
\right\|_1
+
\left\|
\mathbf{b}^{\mathrm{sum}}_{k,M}
\right\|_1
+
\epsilon
}.
\label{eq:symmetric_error}
\end{equation}
The triangle inequality gives $E_{\mathrm{sym}}\in[0,1]$, with lower
values indicating closer magnitude agreement.

\paragraph{Cosine similarity.}
Directional agreement is measured by
\begin{equation}
C_{\mathrm{cos}}
=
\frac{
\left\langle
\mathbf{b}^{\mathrm{joint}}_{k,M},
\mathbf{b}^{\mathrm{sum}}_{k,M}
\right\rangle
}{
\left\|
\mathbf{b}^{\mathrm{joint}}_{k,M}
\right\|_2
\left\|
\mathbf{b}^{\mathrm{sum}}_{k,M}
\right\|_2
+
\epsilon
}.
\label{eq:additivity_cosine}
\end{equation}
This metric complements symmetric error by measuring vector alignment.

\paragraph{Spearman rank correlation.}
We measure global ranking agreement using
\begin{equation}
\rho_{\mathrm{sp}}
=
\operatorname{Spearman}
\left(
\mathbf{b}^{\mathrm{joint}}_{k,M},
\mathbf{b}^{\mathrm{sum}}_{k,M}
\right).
\label{eq:additivity_spearman}
\end{equation}

\paragraph{Top-$k$ overlap.}
To compare highly ranked response positions, we set
\begin{equation}
k_k
=
\max
\left(
1,
\operatorname{round}
\left(
0.25|\mathcal{J}_k|
\right)
\right).
\end{equation}
The overlap is
\begin{equation}
O_k
=
\frac{
\left|
\operatorname{TopK}
\left(
\mathbf{b}^{\mathrm{joint}}_{k,M}
\right)
\cap
\operatorname{TopK}
\left(
\mathbf{b}^{\mathrm{sum}}_{k,M}
\right)
\right|
}{
k_k
}.
\label{eq:additivity_topk}
\end{equation}
The expected overlap fraction under independent random rankings is
\begin{equation}
q_k
=
\frac{k_k}{|\mathcal{J}_k|}.
\end{equation}
We account for this baseline using
\begin{equation}
O_k^{\mathrm{norm}}
=
\frac{
O_k-q_k
}{
1-q_k
},
\label{eq:additivity_topk_norm}
\end{equation}
where zero indicates chance-level overlap and one indicates identical
selections.

\paragraph{Weighted sign agreement.}
We weight sign agreement by the magnitude of the exact joint influence:
\begin{equation}
\begin{split}
A_{\mathrm{sign}}
&= \frac{1}{\sum_{j\in\mathcal{J}_k} |\beta^{\mathrm{joint}}_{k,M,j}| + \epsilon} \sum_{j\in\mathcal{J}_k} |\beta^{\mathrm{joint}}_{k,M,j}| \\
&\quad \times \mathbb{I} \left[ \operatorname{sign}\left(\beta^{\mathrm{joint}}_{k,M,j}\right) = \operatorname{sign}\left(\beta^{\mathrm{sum}}_{k,M,j}\right) \right].
\end{split}
\label{eq:weighted_sign}
\end{equation}
This emphasizes positions with substantial joint influence.

\paragraph{Aggregation and uncertainty.}
Metrics are computed for each example--subset pair. We report means
and 95\% confidence intervals from an example-level cluster bootstrap,
resampling the 200 source examples with replacement and retaining all
perturbation repeats for each sampled example. Zero-variance or
zero-norm vectors are excluded from the corresponding correlation or
cosine metric, respectively, and valid sample counts are reported.

\paragraph{Experimental protocol.}
\label{app:additivity_protocol}
We sample 200 MetaMathQA examples from the SFT split using seed 42,
fixing their identities before measurement. Experiments use the same
approximately 5\% warmup checkpoint and validation direction as BRIDGE, with
gradients restricted to the final four Transformer layers. Each example
includes at most 128 eligible prompt positions and 128 eligible response
positions; BOS is excluded from prompt perturbations, and EOS from
response evaluation. Single-token influences in
Equation~\ref{eq:single_exact_influence} are computed once and reused,
while joint influence is recomputed for each perturbation subset.

\paragraph{Controlled perturbation size.}
\label{app:additivity_local}
We evaluate
\begin{equation}
|M|
\in
\{2,4,8,16,32\}.
\end{equation}
For $|M|=1$, joint and single-token influences coincide:
\begin{equation}
\beta^{\mathrm{joint}}_{k,\{i\},j}
=
\beta_{k,i,j}
\end{equation}
by definition. For each example and evaluated size, we sample three
independent prompt subsets using deterministic seeds fixed before
measurement.

\begin{figure*}[t]
\centering
\includegraphics[width=\textwidth]{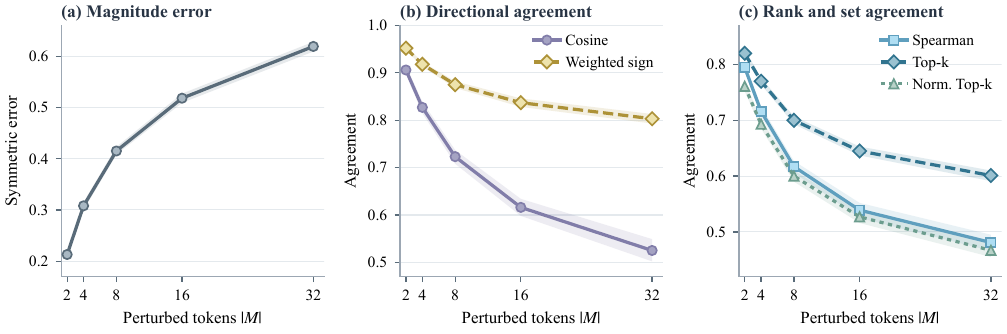}
\vspace{-0.25in}
\caption{
Finite-difference additivity across joint perturbation sizes. Curves
show means over 200 MetaMathQA examples and three deterministic
perturbation subsets per example; shaded bands denote 95\%
example-level cluster-bootstrap confidence intervals.
At $|M|=32$, 198 examples have sufficient eligible prompt positions
($n=198$). Lower symmetric error and higher agreement indicate
greater fidelity.
}

\label{fig:additivity_local}
\end{figure*}

\paragraph{Results.}
At $|M|=2$, cosine similarity is $0.906$, Spearman correlation is
$0.795$, Top-$k$ overlap is $0.820$, and weighted sign agreement is
$0.952$ (Figure~\ref{fig:additivity_local}). At $|M|=32$, symmetric
error is $0.619$, compared with $0.213$ at $|M|=2$, while cosine
similarity and Spearman correlation are $0.525$ and $0.481$,
respectively. Top-$k$ overlap remains $0.601$, with chance-normalized
overlap of $0.467$. These results show that magnitude agreement depends
on perturbation size, while individually measured effects retain
information about highly ranked response positions across the tested
range.

\paragraph{Compression-scale stress test.}
\label{app:additivity_stress}
We also parameterize perturbation size by prompt retention:
\begin{equation}
\rho_p
\in
\{0.9,0.7,0.5,0.3,0.1\},
\end{equation}
replacing the complementary fraction of eligible prompt positions
using deterministically sampled random subsets. This protocol evaluates
joint context changes at the scale of prompt compression.

\begin{table*}[t]
\centering
\small
\setlength{\tabcolsep}{4.5pt}
\resizebox{\textwidth}{!}{%
\begin{tabular}{ccccccc}
\toprule
\textbf{Prompt Retention}
&
\textbf{Sym. Err.} $\downarrow$
&
\textbf{Cosine} $\uparrow$
&
\textbf{Spearman} $\uparrow$
&
\textbf{Top-$k$} $\uparrow$
&
\textbf{Norm. Top-$k$} $\uparrow$
&
\textbf{Weighted Sign} $\uparrow$
\\
\midrule
90\%
& 0.392 (0.379--0.405)
& 0.741 (0.718--0.763)
& 0.645 (0.629--0.661)
& 0.720 (0.709--0.730)
& 0.626 (0.612--0.640)
& 0.891 (0.882--0.900)
\\
70\%
& 0.548 (0.537--0.559)
& 0.590 (0.566--0.614)
& 0.522 (0.506--0.538)
& 0.637 (0.627--0.647)
& 0.516 (0.502--0.530)
& 0.822 (0.810--0.834)
\\
50\%
& 0.617 (0.607--0.627)
& 0.531 (0.507--0.554)
& 0.490 (0.473--0.507)
& 0.606 (0.594--0.617)
& 0.474 (0.458--0.489)
& 0.808 (0.798--0.819)
\\
30\%
& 0.674 (0.665--0.682)
& 0.493 (0.468--0.519)
& 0.459 (0.441--0.477)
& 0.579 (0.567--0.591)
& 0.439 (0.423--0.455)
& 0.786 (0.774--0.798)
\\
10\%
& 0.711 (0.703--0.720)
& 0.458 (0.431--0.484)
& 0.432 (0.414--0.450)
& 0.550 (0.538--0.561)
& 0.399 (0.384--0.415)
& 0.770 (0.757--0.782)
\\
\bottomrule
\end{tabular}%
}
\vspace{-0.1in}
\caption{
Finite-difference additivity across prompt retention ratios.
Lower retention corresponds to a larger jointly perturbed subset.
Parentheses denote 95\% example-level cluster-bootstrap confidence
intervals.
}
\label{tab:additivity_stress}
\vspace{-0.1in}
\end{table*}

\paragraph{Results.}
Table~\ref{tab:additivity_stress} shows the same dependence on
perturbation scale. From $90\%$ to $10\%$ retention, symmetric error
changes from $0.392$ to $0.711$, cosine similarity from $0.741$ to
$0.458$, and Spearman correlation from $0.645$ to $0.432$.
Top-$k$ overlap remains above chance throughout, ranging from $0.720$
to $0.550$, while weighted sign agreement ranges from $0.891$ to
$0.770$. Thus, single-token finite differences retain selection-relevant
ranking and sign information under substantial joint context changes.

\paragraph{Interpretation.}
\label{app:additivity_discussion}
The finite-difference experiments characterize \emph{exact} token
influence, while the state-recomputed comparison directly evaluates
surrogate-based response selection. Together, they show that token
influence varies with the retained context and that token-level scores
provide useful selection structure. The strong reference-utility
retention after physical prompt compaction further supports the BRIDGE
strategy of updating conditional selections through aggregation of
precomputed interactions.

\section{Structural Projection and Supervision Coverage}
\label{app:structural_projection}
Complementing Section~\ref{subsec:structure_projection}, we describe
budgeted structural projection, evaluate supervision coverage, and examine
the effect of an additional prompt update after projection.

\subsection{Projection Rules and Coverage Diagnostics}
\label{app:closure_diagnostics}

\paragraph{Projection details.}
Fixed boundary positions $F_{p,k}$ and $F_{r,k}$ include retained BOS and
supervised EOS, respectively; these positions are excluded from eligible
selection budgets.
Closure holds the selected prompt fixed and uses the final conditional
response scores $s_{k,j}=\Delta_{r,k}(j\mid S_{p,k}^{(T)})$.
Tokenizer-aligned, task-specific boundaries define non-overlapping units.
Base units longer than 96 tokens are split, while units shorter than
4 tokens are merged with a neighboring unit when available. Overlapping
tagged regions are merged before selection, so protected units may exceed
the base-unit length limit. Code and instruction boundaries are described
in Appendices~\ref{app:code_generation}
and~\ref{app:instruction_following}.

For each unit $u$, the core $u_{\mathrm{core}}$ excludes trivial
separators and special tokens where possible; otherwise, it includes all
positions in $u$. The unit score is
\[
q_k(u)=
\frac{\sum_{j\in u_{\mathrm{core}}}\max\{s_{k,j},-\tau\}}
{\left(|u_{\mathrm{core}}|+\epsilon\right)^{\alpha}}.
\]
We use $\tau=0.05$, $\alpha=0.5$, and $\epsilon=10^{-12}$.
In Eq.~\eqref{eq:bridge_unit_utility}, the core and unit-level stabilizer
are denoted by $u^{\mathrm{core}}$ and $\epsilon_q$, respectively.
Raw conditional scores are clipped before aggregation, without
per-example renormalization. Clipping limits the contribution of negative
scores within a unit, while length normalization moderates preference
for longer units. The budget cost $|u|$ counts all supervised positions,
including those excluded from scoring.

Recognized boxed-answer and final-answer units are mandatory. When a
mandatory unit exceeds the available budget, its highest-priority core
positions are retained as capacity permits. Dynamic programming solves
the remaining unit-selection objective in Eq.~\eqref{eq:closure_projection},
breaking utility ties by structural
priority, retained positive utility, and position, in that order.
For residual capacity above 2048, selection uses greedy utility density.
Budget completion repacks whole units or adds individual positions.
When token--text alignment cannot be verified, the original token-level
mask is restored.

\paragraph{Controls and diagnostics.}
We compare four rules under the same response-budget cap:
\emph{Token Random} samples individual positions,
\emph{Span Random} samples contiguous spans,
\emph{Score-only} selects the highest-scoring positions, and
\emph{Score+Closure} applies structural projection.
All variants share the selected prompt and final training procedure.
Table~\ref{tab:closure_diagnostics} reports retained pre-projection
utility (Util.), response-budget utilization (Budget), and the fraction
of examples with complete final-answer supervision (Ans.).

\begin{table}[t]
\centering
\small
\setlength{\tabcolsep}{4.5pt}
\resizebox{\columnwidth}{!}{%
\begin{tabular}{lccccc}
\toprule
\textbf{Rule}
&
\textbf{Util.} $\uparrow$
&
\textbf{Budget} $\uparrow$
&
\textbf{Ans.} $\uparrow$
&
\textbf{MATH}
&
\textbf{General Avg.}
\\
\midrule
Token Random & 0.75 & 1.00 & 0.75 & 11.48 & 46.08 \\
Span Random & 0.77 & 0.98 & 0.37 & 12.63 & 46.05 \\
Score-only & 1.00 & 1.00 & 0.95 & 15.20 & 46.53 \\
Score+Closure & 0.83 & 0.97 & 1.00 & 15.96 & 46.92 \\
\bottomrule
\end{tabular}
}
\vspace{-0.1in}
\caption{
Structural-projection ablation on Llama-3.2-3B with a shared
response-supervision budget cap.
}
\label{tab:closure_diagnostics}
\vspace{-0.1in}
\end{table}

Span Random achieves higher MATH accuracy than Token Random ($12.63$ vs.\ $11.48$), and both score-based variants provide further gains. Compared with Score-only, Score+Closure increases MATH accuracy from $15.20$ to $15.96$, General Avg.\ from $46.53$ to $46.92$, and complete final-answer coverage from $0.95$ to $1.00$. It retains $0.83$ of the pre-projection utility and uses $97\%$ of the target budget. These results support structural projection for improving supervision coverage and downstream performance within the response budget.

\begin{figure}[t]
\centering
\includegraphics[width=\columnwidth]{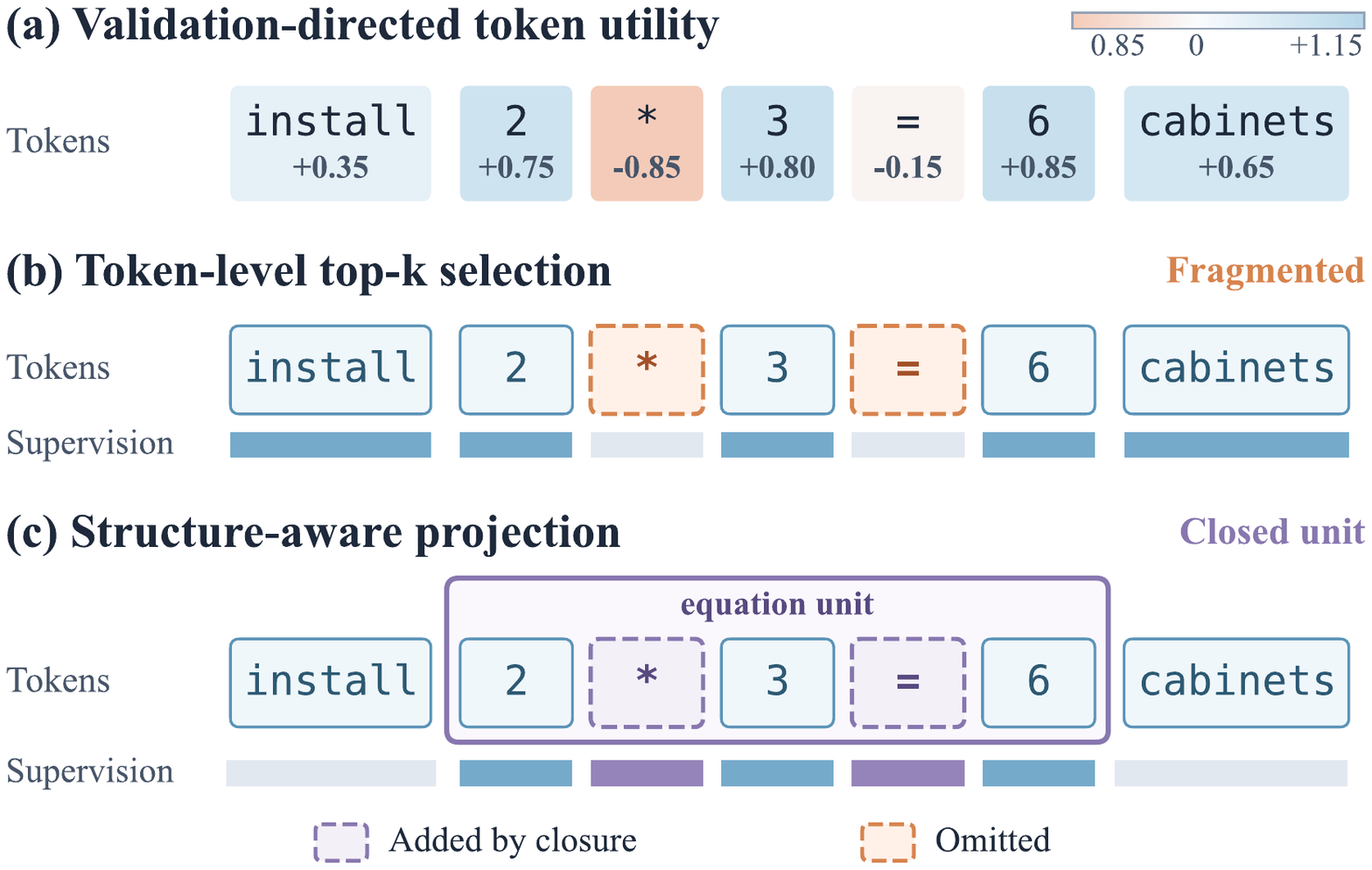}
\vspace{-0.4in}
\caption{Effect of structure-aware projection.}
\label{fig:structure_projection}
\vspace{-0.1in}
\end{figure}

Figure~\ref{fig:structure_projection} illustrates how structure-aware
projection changes response supervision.

\subsection{Post-Closure Prompt Reconditioning}
\label{app:post_closure_refresh}
Closure changes the response supervision set from \(S_r^{(T)}\) to
\(S_r^{\mathrm{cl}}\) within the response-budget cap, while holding
\(S_p^{(T)}\) fixed. We evaluate the effect of conditioning prompt
selection on this projected response set.

Starting from the standard BRIDGE+Closure solution, we perform one
additional prompt update:
\[
S_p^{\mathrm{ref}}
=
\operatorname{Top}_{B_p}
\left\{
\Delta_p(i\mid S_r^{\mathrm{cl}})
\right\}_{i\in P},
\]
with the projected response mask \(S_r^{\mathrm{cl}}\) held fixed.
This isolates post-projection prompt reconditioning.

The refreshed prompt subset has an overlap of \(0.98\) with the
original subset, and downstream performance remains essentially
unchanged (Post-Closure refresh in
Table~\ref{tab:design_analysis}). These results support applying Closure
after alternating selection while retaining the selected prompt.

\section{Computational Cost Breakdown}
\label{app:efficiency_details}
This section supplements Section~\ref{subsec:efficiency} and
Figure~\ref{fig:efficiency} with a breakdown of scoring costs,
throughput, GPU-hours, and stage-specific memory usage.

The total training cost is
\begin{equation}
C_{\mathrm{total}}
=
C_{\mathrm{warmup}}
+
C_{\mathrm{score/select}}
+
C_{\mathrm{final\ SFT}},
\label{eq:appendix_total_cost}
\end{equation}
following the accounting in Section~\ref{subsec:efficiency}.
Warmup and final SFT use standard instruction-tuning procedures.
The BRIDGE-specific scoring and selection overhead consists primarily of
\begin{equation}
C_{\mathrm{extra}}
=
C_{\mathrm{directional}}
+
C_{\mathrm{attention}}
+
C_{\mathrm{selection}}.
\end{equation}
Scoring requires two forward passes per example: one for attention
capture and one for validation-directed scoring, with gradients
restricted to the final four Transformer layers. Interaction-score
assembly and alternating selection follow these passes.
Scoring and selection are performed once per example before final SFT.

Table~\ref{tab:efficiency_breakdown} reports throughput and GPU-hours
under identical data, batch, and optimizer settings.
Figure~\ref{fig:efficiency} compares total cost and peak memory.
All times are normalized to A100-equivalent GPU-hours.

\begin{table}[t]
\centering
\small
\setlength{\tabcolsep}{3.5pt}

\resizebox{\columnwidth}{!}{%
\begin{tabular}{lcccc}
\toprule
\textbf{Method} &
\makecell[c]{\textbf{Throughput}\\\textbf{(tok/s)}} &
\makecell[c]{\textbf{Pre-SFT}\\\textbf{Overhead (GPU-h)}} &
\makecell[c]{\textbf{Final SFT}\\\textbf{(GPU-h)}} &
\makecell[c]{\textbf{Total}\\\textbf{(GPU-h)}} \\
\midrule
Full SFT       & 3.10k & --   & 2.29 & 2.29 \\
Random         & 3.31k & --   & 2.15 & 2.15 \\
RHO-1          & 3.04k & 1.00 & 2.34 & 3.34 \\
LLMLingua-2    & 3.02k & 0.91 & 2.35 & 3.26 \\
ssToken        & 2.76k & 0.28 & 2.57 & 2.85 \\
TokenSeek      & 2.77k & 1.55 & 2.56 & 4.11 \\
Token Cleaning & 3.17k & 1.39 & 2.24 & 3.63 \\
LLM+TC (Ind.)  & 3.05k & 2.15 & 2.33 & 4.48 \\
\rowcolor{oursrow}
\textbf{BRIDGE} & 3.32k & 1.45 & 2.14 & 3.59 \\
\bottomrule
\end{tabular}
}
\vspace{-0.1in}
\caption{
Throughput and GPU-hours on Llama-3.2-3B, normalized to one A100-80GB
GPU. Throughput counts original training tokens per second of final SFT.
Pre-SFT overhead includes warmup, preprocessing, scoring, and selection;
online selection costs are included in final SFT.
}
\label{tab:efficiency_breakdown}
\vspace{-0.1in}
\end{table}

BRIDGE achieves the highest final SFT throughput
($3.32\mathrm{k}$ tok/s) and the lowest final SFT time ($2.14$ GPU-h)
among the compared methods. Prompt compaction reduces executed sequence
length while preserving the complete response as causal history.
The pre-SFT overhead of BRIDGE ($1.45$ GPU-h) is comparable to that of
Token Cleaning ($1.39$ GPU-h) and TokenSeek ($1.55$ GPU-h), and lower
than that of independent LLM+TC composition ($2.15$ GPU-h).
Its total cost is $3.59$ GPU-h, compared with $3.63$ for Token Cleaning,
$4.11$ for TokenSeek, and $4.48$ for LLM+TC.

During scoring, attention rows and directional gradients are materialized
one example at a time, making sequence length the primary determinant
of the per-worker memory footprint. During final SFT, prompt compaction
shortens executed sequences and reduces peak activation memory.

\end{document}